\documentclass[11pt]{article}

\usepackage[preprint]{acl}

\usepackage{times}
\usepackage{latexsym}
\usepackage{booktabs}
\usepackage{makecell}    
\usepackage{array}       
\usepackage[flushleft]{threeparttable} 
\usepackage{enumitem}
\usepackage{multirow}
\usepackage{subfig}
\usepackage{pifont}
\usepackage{xcolor}
\usepackage[most]{tcolorbox}
\definecolor{promptbox_color}{HTML}{9DC8C8}
\definecolor{promptbox_color_2}{HTML}{a5d296}
\usepackage[table]{xcolor}  
\usepackage{colortbl}       
\usepackage[T1]{fontenc}

\usepackage[utf8]{inputenc}

\usepackage{microtype}

\usepackage{inconsolata}
\usepackage[normalem]{ulem}

\usepackage{graphicx}
\usepackage{amsmath} 
\usepackage{amssymb}
\usepackage{tcolorbox}

\title{The Imperfective Paradox in Large Language Models}

\author{Bolei Ma\thanks{Work done while visiting The University of Tokyo.} \\
  LMU Munich \& MCML\\
  \texttt{bolei.ma@lmu.de} 
  \\\And
  Yusuke Miyao \\
  The University of Tokyo \\
  \texttt{yusuke@is.s.u-tokyo.ac.jp} \\
  }

\begin{document}
\maketitle
\begin{abstract}
Do Large Language Models (LLMs) genuinely grasp the compositional semantics of events, or do they rely on surface-level probabilistic heuristics? We investigate the \textbf{Imperfective Paradox}, a logical phenomenon where the past progressive aspect entails event realization for activities (e.g., \textit{running} $\to$ \textit{ran}) but not for accomplishments (e.g., \textit{building} $\nrightarrow$ \textit{built}). We introduce \textsc{ImperfectiveNLI}, a diagnostic dataset designed to probe this distinction across diverse semantic classes. Evaluating state-of-the-art open-weight models, we uncover a pervasive \textbf{Teleological Bias}: models systematically hallucinate completion for goal-oriented events, even overriding explicit textual cancellation. 
Prompting interventions partially reduce this bias but trigger a calibration crisis, causing models to incorrectly reject valid entailments for atelic verbs. 
Representational analyses further show that while internal embeddings often distinguish progressive from simple past forms, inference decisions are dominated by strong priors about goal attainment. Taken together, our findings indicate that these current open-weight LLMs operate as predictive narrative engines rather than faithful logical reasoners, and that resolving aspectual inference requires moving beyond prompting toward structurally grounded alignment.\footnote{The data and code are available at: 
\url{https://github.com/boleima/ImperfectiveParadox}.}
\end{abstract}

\section{Introduction}
Large Language Models (LLMs) have demonstrated remarkable proficiency in natural language inference (NLI) and complex reasoning tasks \citep{openai2023gpt4, touvron2023llama2openfoundation}. Despite these advancements, questions remain regarding the robustness of their reasoning: do models genuinely grasp the compositional semantics of events, or do they rely on surface-level probabilistic heuristics \citep{bender2021dangers, dziri2024faith}? This distinction is critical when processing temporal events. A faithful reasoning agent must distinguish between an action intended (or in progress) and an action realized.

Consider the following sentence: ``\textit{The carpenter was building a gazebo}'', as shown in Table \ref{tab:intro_paradox_single}. Logically, this description of an ongoing process does not entail that the gazebo was ever finished. The construction could have been abandoned or interrupted. This lack of entailment for telic (goal-oriented) events is known in formal semantics as the \textbf{Imperfective Paradox} \citep{Dowty1977, dowty1979word}. In contrast, for atelic verbs like ``\textit{run}'', the process itself constitutes the event \cite{vendler1957,bennett1978toward,Smith1997}: ``\textit{The boy was running}'' entails ``\textit{The boy ran}''.

This phenomenon is grounded in the theory of \textbf{(Lexical) Aspect}, which categorizes events by their telicity (boundedness) \citep{vendler1957, comrie1976aspect, leiss1992}. Unlike atelic activities where the process implies the event (e.g., running), telic accomplishments possess a distinct culmination point that separates the process from the result \citep{moens-steedman-1988-temporal}. Therefore, the progressive aspect acts as a logical filter: it entails realization for activities but suspends it for accomplishments. A robust reasoner must respect this structural boundary, distinguishing between the occurrence of a process and the attainment of a goal.

\begin{table}[t]
    \centering
    \small
    \setlength\tabcolsep{3pt}
    \begin{tabular}{l p{5cm} c}
        \toprule
        \textbf{Class} & \textbf{Examples (Premise $\to$ Hypothesis)} & \textbf{Entail?} \\
        \midrule
        \multirow{2}{*}{\textbf{Telic}} 
            & \textbf{P:} The carpenter \textit{was building} a gazebo. 
            & \multirow{2}{*}{\textcolor{red}{\textbf{\ding{55} No}}} \\
            & \textbf{H:} The carpenter \textit{built} a gazebo. & \\
        \midrule
        \multirow{2}{*}{\textbf{Atelic}} 
            & \textbf{P:} The boy \textit{was running} in the park. 
            & \multirow{2}{*}{\textcolor{green}{\textbf{\ding{51} Yes}}} \\
            & \textbf{H:} The boy \textit{ran} in the park. & \\
        \bottomrule
    \end{tabular}
    \caption{Examples for the Imperfective Paradox. Telic verbs do not entail completion in the progressive form.}
    \label{tab:intro_paradox_single}
\end{table}

While humans intuitively navigate these aspectual distinctions, we hypothesize that LLMs exhibit a \textbf{Teleological Bias}: a tendency to assume that goal-directed actions inevitably lead to their successful completion. This bias likely stems from the ``reporting bias'' in pre-training corpora \citep{gordon-vandurme-2013-reporting}, where narratives typically mention goals only when they are relevant to the outcome, often implying success \citep{Grice1975LogicAC}. Consequently, models may hallucinate event completion based on statistical likelihood rather than logical entailment, effectively conflating the intent with the result \cite{Farquhar2024}.

Prior work in NLP has largely focused on aspect classification or modeling, training models to label verbs as stative, dynamic, telic, or atelic (\citealp[see][]{friedrich-etal-2023-kind}). However, high classification accuracy does not guarantee robust inference. It remains unexplored whether generative models can apply these aspectual properties to resist fallacious entailments in zero-shot reasoning scenarios.

To bridge this gap, we introduce \textsc{ImperfectiveNLI}, a diagnostic dataset to probe the Imperfective Paradox. We construct minimal pairs across four logical conditions, contrasting interrupted and ambiguous contexts for both telic and atelic verbs. 
We evaluate state-of-the-art open-weight models and uncover following critical insights:

\begin{enumerate}[leftmargin=*]
    \item[1)] \textbf{Teleological Bias Dominates Model Inference:} Models exhibit a pervasive tendency to hallucinate completion for goal-oriented events. Crucially, this bias is robust enough to override explicit textual negation (e.g., predicting \textit{completion} even when an action is interrupted), revealing a fundamental failure in \textit{contextual fidelity}.
    \item[2)] \textbf{Prompting Interventions Trigger a Calibration Crisis:} Strategies like Chain-of-Thought or forced counterfactual non-endpoints can reduce hallucinated completions, but often over-correct, causing models to incorrectly reject valid entailments for atelic activities. Models struggle to adapt reasoning dynamically based on event type.
    \item[3)] \textbf{Representation and Reasoning Are Dissociated:} We find that the bias is not representational; embeddings successfully distinguish \textit{process} (progressive form) from \textit{result} (simple past form). Instead, inference is dominated by prior expectations about goal completion. 
    The failure is likely driven by predictive priors during decoding, rather than an inherent inability of Transformer architectures to encode the aspectual distinction.
\end{enumerate}

Ultimately, our findings underscore that current LLMs struggle to resolve the Imperfective Paradox, operating as probabilistic predictors of narrative outcomes rather than faithful logical reasoners. 
Bridging this gap requires not just better prompting, but a deeper alignment between model behavior and the formal logic of event semantics.

\section{Background}

\subsection{Vendler’s Aspectual Classes: Telicity}

\citet{vendler1957} categorized verbs into four aspectual classes (activity, accomplishment, achievement, state) based on their temporal properties. In this study, we focus on the distinction between Activities and Accomplishments, defined by the property of \textit{telicity}, whether an event involves an inherent endpoint. Telicity often correlates with perfective interpretations, which present an eventuality from a holistic perspective \cite{Filip2008-FILEAM}.

\begin{itemize}[leftmargin=*]
    \item \textbf{Activities (Atelic)}: Events with no inherent endpoint (e.g., ``\textit{run}'', ``\textit{swim}'', ``\textit{push a cart}''). They are homogeneous; any part of the process counts as an instance of the same event.
    \item \textbf{Accomplishments (Telic)}: Events that proceed toward a specific terminus or goal (e.g., ``\textit{build a house}'', ``\textit{write a letter}''). Such events are only completed and receive canonical perfective interpretations, when the endpoint is reached.
\end{itemize}

\subsection{The Imperfective Paradox}
Semantically, in English, the progressive form has imperfective reading, while the simple past form has perfective reading \citep{comrie1976aspect, Smith1997}. 
The Imperfective Paradox arises when we analyze the logical entailment between the progressive form (Imperfective) and the simple past form (Perfective) of these verb classes. Let $\phi$ be an event predicate and $PROG(\phi)$ denote its progressive form. 

For Atelic predicates (Activities), the entailment holds:$$PROG(\text{run}) \models \text{run}$$
If $x$ was running, then it is true that $x$ ran. This is often referred to as the Sub-interval Property: if an activity holds true for an interval, it holds true for any sub-interval \cite{bennett1978toward}.

For Telic predicates (Accomplishments), the entailment fails:$$PROG(\text{build a house}) \not\models \text{build a house}$$
If $x$ was building a house, it is NOT necessarily true that $x$ built a house.

The paradox highlights that the progressive form of an accomplishment describes a process leading to a result, but does not assert the result itself \cite{Parsons1990}. \citet{Dowty1977} formalized this using possible worlds semantics, suggesting that $PROG(\phi)$ is true if $\phi$ completes in all ``inertia worlds'' (worlds where the course of events proceeds without interruption). However, the real world is not necessarily an inertia world and interruptions occur.

\subsection{Related Work: Interpreting and Modeling ``Aspect'' in NLP}

Computational modeling of aspect has evolved from feature-engineering approaches to representation learning \cite{friedrich-etal-2023-kind}. \citet{siegel-mckeown-2000-learning} pioneered the use of supervised machine learning for aspectual classification. Subsequent work extensively explored semi-supervised approaches, combining linguistic and distributional features to predict stativity, duration, and situation entity types, while also establishing essential annotated datasets \cite{friedrich-palmer-2014-automatic, friedrich-pinkal-2015-automatic, friedrich-etal-2016-situation, friedrich-gateva-2017-classification}. Similar methodologies were applied to German verbs \cite{hermes2015automatic} and multilingual morphosyntactic annotation \cite{ramm-etal-2017-annotating}.

Moving toward compositional semantics, \citet{kober-etal-2020-aspectuality} introduced an NLI-based dataset to model aspect in context. Later work showed that transformer models capture some aspectual cues but rely heavily on surface features such as tense and word order \citep{metheniti-etal-2022-time}. More recently, \citet{ma-2024-evaluating} found that large-scale LLMs still struggle with lexical aspect identification. 
Closely related are two studies on the Imperfective Paradox \cite{prus-etal-2024-human, prus-etal-2025-world}, which focus primarily on modeling human judgments of progressive versus simple past forms, treating LLM behavior as a secondary comparison. 
As such, neither study provides a systematic diagnostic evaluation of LLM aspectual inference. Our work directly addresses this gap, offering the first controlled investigation of whether LLMs respect the Imperfective Paradox or instead hallucinate event completion by conflating goal-directed processes with their result states.

\section{Dataset - \textsc{ImperfectiveNLI}}
\label{sec:dataset} 
To evaluate LLMs' ability to reason about the imperfective paradox and teleological aspectual semantics, we introduce \textsc{ImperfectiveNLI}, a diagnostic NLI dataset in English. 

\paragraph{Dataset Construction with Different Verbs.}
The construction of \textsc{ImperfectiveNLI} employs a template-based generation approach to ensure syntactic uniformity while isolating semantic reasoning. We curated a balanced lexicon of verbs based on Vendler’s aspectual classification:

\begin{itemize}[leftmargin=*]
    \item \textbf{100 Telic Verbs (Accomplishments):} Predicates involving a natural endpoint or result state, where the process is distinct from the culmination (e.g., \textit{build, write, fix, paint, bake}).

    \item \textbf{100 Atelic Verbs (Activities):} Predicates describing homogeneous processes without a necessary endpoint, where the process itself constitutes the event (e.g., \textit{run, swim, wander, play, vibrate}).
\end{itemize}

For each verb, we generated context-hypothesis pairs using Gemini-assisted rewriting \cite{geminiteam2025geminifamilyhighlycapable}, followed by strict human verification. The dataset is structured as a controlled minimal-pair experiment across four logical conditions, resulting in a total of $N=400$ examples.

To investigate whether teleological bias varies across different types of goal-directedness, we further annotated the 100 telic verbs into four fine-grained semantic classes (inspired by the theory of \citealp{levin1993}): 
\textit{Change of State} (44), \textit{Creation} (39), \textit{Consumption} (9), and \textit{Motion to Goal} (8). 

We restricted this sub-classification to the telic group because, unlike homogeneous activities, telic predicates inherently encode a distinct \textit{resultant state} \citep{rappaporthovav2010}. This classification categorizes the nature of that result: \textit{Change of State} verbs denote a transformation in the physical or abstract condition of an existing entity (e.g., \textit{melt}, \textit{repair}); \textit{Creation} verbs involve bringing an object into existence (e.g., \textit{build}, \textit{knit}); \textit{Consumption} verbs entail the destruction or complete ingestion of an object (e.g., \textit{eat}, \textit{burn}) \citep{beavers2010}; and \textit{Motion to Goal} verbs describe movement towards a specific spatial endpoint (e.g., \textit{climb}, \textit{cross}) \citep{talmy2000}. This stratification enables us to probe whether LLMs are more prone to hallucinating completion for artifact-creating events, where the ``teleological'' purpose is most salient, compared to simple state changes.

\begin{table*}[t]
\centering
\small
\renewcommand\arraystretch{0.8} 

\resizebox{\textwidth}{!}{
\begin{tabular}{l p{3cm} p{4.9cm} p{3.7cm} c}
\toprule
\textbf{Group} & \textbf{Condition} & \textbf{Premise Example} & \textbf{Hypothesis} & \textbf{Gold Label} \\
\midrule

\multirow{2}{*}{\textbf{A}} 
& \textbf{Interrupted Accomp.} \newline {\footnotesize (Telic + Cancel)} 
& The carpenter was building a gazebo, \textit{but a storm destroyed the frame before the roof was on}. 
& The carpenter built a gazebo. 
& \multirow{1}{*}{\shortstack{Contradiction\\(\textcolor{red}{\textbf{\ding{55}}} False)}} \\ 
\addlinespace[1.5ex] 

\multirow{2}{*}{\textbf{B}} 
& \textbf{Interrupted Activity} \newline {\footnotesize (Atelic + Stop)} 
& The athletes were running on the track, \textit{but it started to hail}. 
& The athletes ran on the track. 
& \multirow{1}{*}{\shortstack{Entailment\\(\textcolor{green}{\textbf{\ding{51}}} True)}} \\
\addlinespace[1.5ex]

\multirow{2}{*}{\textbf{C}} 
& \textbf{Ambiguous Accomp.} \newline {\footnotesize (Telic + Process)} 
& The carpenter was building a gazebo. 
& The carpenter built a gazebo. 
& \multirow{1}{*}{\shortstack{Neutral\\(\textcolor{brown}{\textbf{?}} Unknown)}} \\
\addlinespace[1.5ex]

\multirow{2}{*}{\textbf{D}} 
& \textbf{Ambiguous Activity} \newline {\footnotesize (Atelic + Process)} 
& The athletes were running on the track. 
& The athletes ran on the track. 
& \multirow{1}{*}{\shortstack{Entailment\\(\textcolor{green}{\textbf{\ding{51}}} True)}} \\

\bottomrule
\end{tabular}
}
\caption{Examples from the \textsc{ImperfectiveNLI} dataset. Group C is the critical probe, where the imperfective paradox implies a \textit{Neutral} relation, but teleological bias may lead models to predict \textit{Entailment}.}
\label{tab:dataset_examples}
\end{table*}

\paragraph{Imperfective Logical Conditions.} 
The core contribution of \textsc{ImperfectiveNLI} is its minimal-pair design, which can be viewed as a $2 \times 2$ matrix crossing \textbf{Verb Telicity} (Telic vs. Atelic) with \textbf{Contextual Information} (Interrupted vs. Ambiguous), described in the following groups A-D. Table \ref{tab:dataset_examples} illustrates these conditions with examples.

\begin{tcolorbox}[
    colframe=promptbox_color,      
    colback=white,                 
    coltitle=black,                
    colbacktitle=promptbox_color,  
    rounded corners,
    arc=1.5mm, 
    enhanced,                    
    boxrule=0.6mm,                
    frame style={solid},          
    fonttitle=\bfseries,          
    fontupper=\small,
    title=\small {Group A: Interrupted Accomplishment.},
    boxrule=0.8pt, left=2pt, right=2pt, top=2pt, bottom=2pt
]

    This group serves as a counterfactual check. The premise describes a telic action in progress but includes an explicit cancellation clause (e.g., \textit{``...but a storm destroyed the frame before the roof was on.''}). This tests whether models can attend to explicit interruption of action and the negation of the result. The logical relation is \textbf{Contradiction}.
\end{tcolorbox}

\begin{tcolorbox}[
    colframe=promptbox_color,      
    colback=white,                 
    coltitle=black,                
    colbacktitle=promptbox_color,  
    rounded corners,
    arc=1.5mm, 
    enhanced,                    
    boxrule=0.6mm,                
    frame style={solid},          
    fonttitle=\bfseries,          
    fontupper=\small,
    title=\small{Group B: Interrupted Activity.},
    boxrule=0.8pt, left=2pt, right=2pt, top=2pt, bottom=2pt
]
    This group controls for aspectual class properties. Due to the \textit{sub-interval property} of atelic verbs, stopping an activity does not negate its occurrence (i.e., stopping running implies one has run). Even with an explicit interruption, the premise entails the hypothesis. The logical relation is \textbf{Entailment}.
\end{tcolorbox}

\begin{tcolorbox}[
    colframe=promptbox_color,      
    colback=white,                 
    coltitle=black,                
    colbacktitle=promptbox_color,  
    rounded corners,
    arc=1.5mm, 
    enhanced,                    
    boxrule=0.6mm,                
    frame style={solid},          
    fonttitle=\bfseries,          
    fontupper=\small,
    title=\small{Group C: Ambiguous Accomplishment.},
    boxrule=0.8pt, left=2pt, right=2pt, top=2pt, bottom=2pt
]
This condition is the primary testbed for the Imperfective Paradox. Structurally, Group C is an ablation of Group A, where the interruption clause is removed, leaving only the progressive form of a telic verb. Semantically, the outcome is under-specified: the action was in progress, but completion is not known. The logical relation is \textbf{Neutral}. This group probes whether models hallucinate completion in the absence of explicit evidence.

\end{tcolorbox}

\begin{tcolorbox}[
    colframe=promptbox_color,      
    colback=white,                 
    coltitle=black,                
    colbacktitle=promptbox_color,  
    rounded corners,
    arc=1.5mm, 
    enhanced,                    
    boxrule=0.6mm,                
    frame style={solid},          
    fonttitle=\bfseries,          
    fontupper=\small,
    title=\small{Group D: Ambiguous Activity.},
    boxrule=0.8pt, left=2pt, right=2pt, top=2pt, bottom=2pt
]
Analogous to Group C, this condition is an ablation of Group B, retaining only the progressive form of an activity. Unlike Group C, the entailment holds here because the atelic process itself validates the event. This baseline tests whether models possess basic aspectual competence for atelic verbs. The logical relation is \textbf{Entailment}.
\end{tcolorbox}

While standard NLI tasks utilize the labels \textit{Entailment}, \textit{Contradiction}, and \textit{Neutral} (\citealp[e.g.,][]{bowman-etal-2015-large}), we map these to the natural language truth values \textbf{``True''}, \textbf{``False''}, and \textbf{``Unknown''} respectively for our experiments. 

\paragraph{Quality Assurance.} 
To ensure the validity and naturalness of the generated data, we recruited three native English speakers as evaluators via \textit{Prolific}\footnote{\url{https://www.prolific.com/}}. They assessed the sentences across three linguistic dimensions, Grammar, Fluency, and Adequacy (details in Appendix \ref{app:human}), and were compensated at a rate of £9/hour. Beyond quantitative ratings, evaluators provided corrections for any unnatural or unclear instances. Discrepancies and conflicting revisions were resolved through manual verification, ensuring quality in the final dataset.

\section{Experimental Settings}

\subsection{Models} 
We experimented with 7 instruction-tuned open-weight models with their parameter around 7-9B: \texttt{Llama-3.1-8B-Instruct} \cite{grattafiori2024llama3herdmodels}, \texttt{Mistral-7B-Instruct-v0.3} \cite{jiang2023mistral7b}, \texttt{Qwen2.5-7B-Instruct} \cite{qwen2025qwen25technicalreport}, \texttt{deepseek-llm-7b-chat} \cite{deepseekai2024deepseekv3technicalreport}, \texttt{gemma-2-9b-it} \cite{gemmateam2024}, \texttt{glm-4-9b-chat-hf} \cite{glm2024chatglmfamilylargelanguage}, \texttt{Yi-1.5-9B-Chat} \cite{ai2025yiopenfoundationmodels}.

\subsection{Prompts}
To disentangle the model's intrinsic biases from its instruction-following capabilities, we evaluate four distinct prompting strategies. The detailed prompts given to the models are presented in Appendix \ref{app:prompt}. 

\paragraph{(1) Strict Logic Prompt (Zero-Shot).}
Our primary setting employs a simple zero-shot instruction, framing the model as a ``strict logician''. This prompt explicitly prohibits common sense assumptions (e.g., the pragmatic inference that ``building usually implies a house'') and forces the model to rely solely on the provided text. 


\paragraph{(2) Definition-Aware Prompt (DAP).}
To test whether the failure stems from a lack of linguistic \textit{knowledge} or a failure of \textit{application}, we explicitly inject the linguistic definition of the imperfective aspect for both Activity and Accomplishment into the system prompt. If the bias persists even when the rule is provided, it suggests that the teleological prior is robust enough to override in-context instructions.


\paragraph{(3) Chain-of-Thought Prompt (CoT).}
We adapt the standard ``Let's think step by step'' paradigm \citep{wei2022chain}. This setting investigates whether the Teleological Bias is a result of fast, superficial processing. We hypothesize that by forcing the model to articulate the temporal status of the event before predicting the label, it may self-correct the hallucination for Ambiguous Accomplishment.


\paragraph{(4) Counterfactual Prompt (Counterfactual).}
Drawing on \citet{Dowty1977}'s theory that the progressive aspect presupposes an ``inertia world'' (where events proceed normally), this prompt attempts to explicitly break this assumption. We explicitly instruct the model to first generate three plausible real-world scenarios in which the event is interrupted and its endpoint is not reached, before making a judgment.

\subsection{Metrics} We report following metrics:

\paragraph{(1) Group-wise Accuracy (ACC): } We calculate classification accuracy separately for each logical condition (Groups A--D).


\paragraph{(2) Teleological Bias Rate (TBR):} 
This metric specifically quantifies the ``Hallucination of Completion'' in Group C (Ambiguous Accomplishment). Since the Gold Label for Group C is \textit{Unknown}, any prediction of \textit{True} indicates that the model has hallucinated a result. We define TBR as: 
\begin{equation}
\text{TBR}_G = \frac{\sum_{i \in G} \mathbb{I}(\hat{y}_i = \text{True})}{|G|}
\end{equation}
where $G$ is the set of examples in Group C (or Group A), and $\hat{y}_i$ is the model's prediction. A higher TBR indicates a stronger reliance on teleological priors.


\paragraph{(3) Aspectual Awareness Gap ($\Delta_{AA}$):} 
This metric measures the model's capability to distinguish the logical implications of \textit{atelic} processes (which imply realization) from \textit{telic} processes (which do not). As we are mostly interested in Group C, it is calculated as the difference between the accuracy on the Ambiguous Activity baseline (Group D) and the error rate on the Ambiguous Accomplishment probe (Group C):
\begin{equation}
\Delta_{AA} = \text{ACC}_D - \text{TBR}_C
\end{equation}
where $\text{ACC}_D$ represents the correct entailment rate for Activity verbs. A high gap 
indicates ideal aspectual reasoning: the model correctly infers entailment for activities while correctly suspending judgment (low bias) for accomplishments. A low gap 
suggests that the model treats all progressive verbs homogeneously, likely applying a shallow heuristic (e.g., ``mentioning an action implies its completion'') regardless of lexical aspect. 

Note that $\Delta_{AA}$ is most interpretable when $\text{ACC}_D$ is high; a low $\Delta_{AA}$  driven by a low $\text{ACC}_D$ rather than a high $\text{TBR}_C$ reflects a distinct failure mode, namely, an inability to recognize valid entailments for atelic verbs, rather than teleological overgeneralization.

\begin{table*}[t]
\centering
\renewcommand\arraystretch{0.95}
\footnotesize
\setlength{\tabcolsep}{3pt} 
\begin{tabular}{llccccccc}
\toprule
\textbf{Prompt} & \textbf{Model} & \textbf{Acc A} ($\uparrow$) & \textbf{Acc B} ($\uparrow$) & \textbf{Acc C} ($\uparrow$) & \textbf{Acc D} ($\uparrow$) & \textbf{TBR$_A$} ($\downarrow$) & \textbf{TBR$_C$} ($\downarrow$) & \textbf{$\Delta_{AA}$} ($\uparrow$) \\
\midrule
\multirow{7}{*}{\textbf{Zero-shot}} 
&Llama-3.1-8B-Instruct 
& \cellcolor{purple!23.5}0.47 
& \cellcolor{purple!42.5}0.85 
& \cellcolor{purple!1}0.02 
& \cellcolor{purple!49}0.98 
& \cellcolor{purple!25.5}0.51 
& \cellcolor{purple!49}0.98 
& \cellcolor{purple!0}0.00 \\
&Mistral-7B-Instruct-v0.3 
& \cellcolor{purple!18.5}0.37 
& \cellcolor{purple!46}0.92 
& \cellcolor{purple!1}0.02 
& \cellcolor{purple!50}1.00 
& \cellcolor{purple!23}0.46 
& \cellcolor{purple!48.5}0.97 
& \cellcolor{purple!1.5}0.03 \\
&Qwen2.5-7B-Instruct 
& \cellcolor{purple!10}0.20 
& \cellcolor{purple!49}0.98 
& \cellcolor{purple!23.5}0.47 
& \cellcolor{purple!48.5}0.97 
& \cellcolor{purple!27.5}0.55 
& \cellcolor{purple!26.5}0.53 
& \cellcolor{purple!22}0.44 \\
&Yi-1.5-9B-Chat 
& \cellcolor{purple!17.5}0.35 
& \cellcolor{purple!47}0.94 
& \cellcolor{purple!1}0.02 
& \cellcolor{purple!50}1.00 
& \cellcolor{purple!28.5}0.57 
& \cellcolor{purple!49}0.98 
& \cellcolor{purple!1}0.02 \\
&deepseek-llm-7b-chat 
& \cellcolor{purple!2}0.04 
& \cellcolor{purple!44}0.88 
& \cellcolor{purple!0}0.00 
& \cellcolor{purple!50}1.00 
& \cellcolor{purple!23.5}0.47 
& \cellcolor{purple!50}1.00 
& \cellcolor{purple!0}0.00 \\
&gemma-2-9b-it 
& \cellcolor{purple!1.5}0.03 
& \cellcolor{purple!48}0.96 
& \cellcolor{purple!3}0.06 
& \cellcolor{purple!50}1.00 
& \cellcolor{purple!25}0.50 
& \cellcolor{purple!47}0.94 
& \cellcolor{purple!3}0.06 \\
&glm-4-9b-chat-hf 
& \cellcolor{purple!7}0.14 
& \cellcolor{purple!49}0.98 
& \cellcolor{purple!1.5}0.03 
& \cellcolor{purple!50}1.00 
& \cellcolor{purple!37.5}0.75 
& \cellcolor{purple!48.5}0.97 
& \cellcolor{purple!1.5}0.03 \\
\midrule
\multirow{7}{*}{\textbf{DAP}} 
& Llama-3.1-8B-Instruct 
& \cellcolor{purple!30}0.60 
& \cellcolor{purple!47.5}0.95 
& \cellcolor{purple!18}0.36 
& \cellcolor{purple!49.5}0.99 
& \cellcolor{purple!17.5}0.35 
& \cellcolor{purple!22.5}0.45 
& \cellcolor{purple!27}0.54 \\
& Mistral-7B-Instruct-v0.3 
& \cellcolor{purple!27.5}0.55 
& \cellcolor{purple!48}0.96 
& \cellcolor{purple!3}0.06 
& \cellcolor{purple!50}1.00 
& \cellcolor{purple!20.5}0.41 
& \cellcolor{purple!47}0.94 
& \cellcolor{purple!3}0.06 \\
& Qwen2.5-7B-Instruct 
& \cellcolor{purple!17.5}0.35 
& \cellcolor{purple!48.5}0.97 
& \cellcolor{purple!44.5}0.89 
& \cellcolor{purple!36}0.72 
& \cellcolor{purple!15}0.30 
& \cellcolor{purple!4.5}0.09 
& \cellcolor{purple!31.5}0.63 \\
& Yi-1.5-9B-Chat 
& \cellcolor{purple!28.5}0.57 
& \cellcolor{purple!48.5}0.97 
& \cellcolor{purple!10}0.20 
& \cellcolor{purple!50}1.00 
& \cellcolor{purple!18}0.36 
& \cellcolor{purple!40}0.80 
& \cellcolor{purple!10}0.20 \\
& deepseek-llm-7b-chat 
& \cellcolor{purple!6}0.12 
& \cellcolor{purple!46}0.92 
& \cellcolor{purple!0}0.00 
& \cellcolor{purple!50}1.00 
& \cellcolor{purple!20}0.40 
& \cellcolor{purple!50}1.00 
& \cellcolor{purple!0}0.00 \\
& gemma-2-9b-it 
& \cellcolor{purple!4}0.08 
& \cellcolor{purple!45}0.90 
& \cellcolor{purple!30}0.60 
& \cellcolor{purple!44}0.88 
& \cellcolor{purple!15.5}0.31 
& \cellcolor{purple!20}0.40 
& \cellcolor{purple!24}0.48 \\
& glm-4-9b-chat-hf 
& \cellcolor{purple!4}0.08 
& \cellcolor{purple!50}1.00 
& \cellcolor{purple!16.5}0.33 
& \cellcolor{purple!49}0.98 
& \cellcolor{purple!35}0.70 
& \cellcolor{purple!33.5}0.67 
& \cellcolor{purple!15.5}0.31 \\
\midrule
\multirow{7}{*}{\textbf{CoT}} 
& Llama-3.1-8B-Instruct 
& \cellcolor{purple!7.5}0.15 
& \cellcolor{purple!19.5}0.39 
& \cellcolor{purple!33.5}0.67 
& \cellcolor{purple!32.5}0.65 
& \cellcolor{purple!5}0.10 
& \cellcolor{purple!16.5}0.33 
& \cellcolor{purple!16}0.32 \\
& Mistral-7B-Instruct-v0.3 
& \cellcolor{purple!11.5}0.23 
& \cellcolor{purple!32.5}0.65 
& \cellcolor{purple!20.5}0.41 
& \cellcolor{purple!33}0.66 
& \cellcolor{purple!3.5}0.07 
& \cellcolor{purple!28}0.56 
& \cellcolor{purple!5}0.10 \\
& Qwen2.5-7B-Instruct 
& \cellcolor{purple!15}0.30 
& \cellcolor{purple!40}0.80 
& \cellcolor{purple!48.5}0.97 
& \cellcolor{purple!18.5}0.37 
& \cellcolor{purple!6.5}0.13 
& \cellcolor{purple!1.5}0.03 
& \cellcolor{purple!17}0.34 \\
& Yi-1.5-9B-Chat 
& \cellcolor{purple!28}0.56 
& \cellcolor{purple!40.5}0.81 
& \cellcolor{purple!42.5}0.85 
& \cellcolor{purple!22}0.44 
& \cellcolor{purple!7}0.14 
& \cellcolor{purple!6.5}0.13 
& \cellcolor{purple!15.5}0.31 \\
& deepseek-llm-7b-chat 
& \cellcolor{purple!2.5}0.05 
& \cellcolor{purple!44}0.88 
& \cellcolor{purple!20}0.40 
& \cellcolor{purple!47}0.94 
& \cellcolor{purple!28}0.56 
& \cellcolor{purple!30}0.60 
& \cellcolor{purple!17}0.34 \\
& gemma-2-9b-it 
& \cellcolor{purple!7.5}0.15 
& \cellcolor{purple!44}0.88 
& \cellcolor{purple!49}0.98 
& \cellcolor{purple!7.5}0.15 
& \cellcolor{purple!13}0.26 
& \cellcolor{purple!1}0.02 
& \cellcolor{purple!6.5}0.13 \\
& glm-4-9b-chat-hf 
& \cellcolor{purple!12}0.24 
& \cellcolor{purple!47.5}0.95 
& \cellcolor{purple!19}0.38 
& \cellcolor{purple!48.5}0.97 
& \cellcolor{purple!27}0.54 
& \cellcolor{purple!31}0.62 
& \cellcolor{purple!17.5}0.35 \\
\midrule
\multirow{7}{*}{\textbf{Counterfactual}} 
& Llama-3.1-8B-Instruct 
& \cellcolor{purple!17.5}0.35 
& \cellcolor{purple!0}0.00 
& \cellcolor{purple!48.5}0.97 
& \cellcolor{purple!0}0.00 
& \cellcolor{purple!0}0.00 
& \cellcolor{purple!0}0.00 
& \cellcolor{purple!0}0.00 \\
& Mistral-7B-Instruct-v0.3 
& \cellcolor{purple!14}0.28 
& \cellcolor{purple!19.5}0.39 
& \cellcolor{purple!43.5}0.87 
& \cellcolor{purple!9.5}0.19 
& \cellcolor{purple!2.5}0.05 
& \cellcolor{purple!5.5}0.11 
& \cellcolor{purple!4}0.08 \\
& Qwen2.5-7B-Instruct 
& \cellcolor{purple!3.5}0.07 
& \cellcolor{purple!28.5}0.57 
& \cellcolor{purple!50}1.00 
& \cellcolor{purple!1}0.02 
& \cellcolor{purple!7}0.14 
& \cellcolor{purple!0}0.00 
& \cellcolor{purple!1}0.02 \\
& Yi-1.5-9B-Chat 
& \cellcolor{purple!3.5}0.07 
& \cellcolor{purple!20.5}0.41 
& \cellcolor{purple!50}1.00 
& \cellcolor{purple!5}0.10 
& \cellcolor{purple!4.5}0.09 
& \cellcolor{purple!0}0.00 
& \cellcolor{purple!5}0.10 \\
& deepseek-llm-7b-chat 
& \cellcolor{purple!12.5}0.25 
& \cellcolor{purple!48.5}0.97 
& \cellcolor{purple!0}0.00 
& \cellcolor{purple!49}0.98 
& \cellcolor{purple!33.5}0.67 
& \cellcolor{purple!50}1.00 
& \cellcolor{purple!0}-0.02 \\
& gemma-2-9b-it 
& \cellcolor{purple!1.5}0.03 
& \cellcolor{purple!0}0.00 
& \cellcolor{purple!50}1.00 
& \cellcolor{purple!0}0.00 
& \cellcolor{purple!0}0.00 
& \cellcolor{purple!0}0.00 
& \cellcolor{purple!0}0.00 \\
& glm-4-9b-chat-hf 
& \cellcolor{purple!11.5}0.23 
& \cellcolor{purple!33}0.66 
& \cellcolor{purple!36.5}0.73 
& \cellcolor{purple!12}0.24 
& \cellcolor{purple!10.5}0.21 
& \cellcolor{purple!10}0.20 
& \cellcolor{purple!2}0.04 \\
\bottomrule
\end{tabular}
\caption{Model performance under Zero-shot, DAP, CoT, and Counterfactual prompting strategies with Acc for A-D Groups, Teleological Bias Rate for Groups A and C (TBR$_A$, TBR$_C$), and Aspectual Awareness Gap ($\Delta_{AA}$).}
\label{tab:aspectual_results}
\end{table*}

\section{Results and Analysis}

In this section, we present the evaluation results on \textsc{ImperfectiveNLI}. We aim to answer four key questions: whether LLMs exhibit teleological bias, how prompting strategies mitigate this bias, how the model scale affects the bias, and how these biases manifest across different semantic categories.

\subsection{The Dominance of Teleological Bias for \textit{``If A Subject was V-ing, then the Subject V-ed''} in Zero-shot}
\label{sec:zero-shot}
The first few rows of Table \ref{tab:aspectual_results} present the performance of seven open-weight models in the zero-shot setting. The results reveal a fundamental inability of current LLMs to decouple the description of a process from the assertion of its result.

\paragraph{$P(\text{Complete} \mid \text{Past Progressive}) \approx$ 1? -- -- The Illusion of Competence.}
At first glance, models appear competent in aspectual reasoning, achieving near-perfect accuracy on Group D for ambiguous Activity (atelic verbs) (e.g., \texttt{Llama-3.1}: $0.98$, \texttt{Mistral}: $1.00$). However, this high performance is deceptive. When contrasted with the catastrophic failure on Group C for ambiguous Accomplishment (telic verbs) (e.g., \texttt{Llama-3.1}: $0.02$), a different picture emerges. 
The near-zero $\Delta_{AA}$ for most models indicates that they do not genuinely understand the \textit{Sub-interval Property} of the atelic verbs \citep{bennett1978toward}. Instead, they likely employ a shallow heuristic: \textit{``If Subject was V-ing, then Subject V-ed''}, regardless of verbs in the Zero-shot setting. This aligns with findings by the NLI work \citet{mccoy-etal-2019-right}, suggesting that the 
models might often rely on lexical overlap strategies rather than deep semantic compositionality.

\paragraph{Contextual Neglect via Semantic Priming.} 
Perhaps the most alarming finding is the resistance to explicit negation in Group A (Interrupted Accomplishment). Despite the textually grounded interruption (e.g., \textit{``...a storm destroyed the frame before the roof was on''}), most models exhibit a stark performance degradation. This stands in sharp contrast to Group B (Interrupted Activity), where models maintain high accuracy. This performance gap is semantically revealing: for activities, an interruption does not negate the event's occurrence (the \textit{sub-interval property} ensures that ``stopped running'' entails ``ran''), whereas for accomplishments, interruption strictly precludes the result.

The failure in Group A suggests that the models' reasoning is driven by \textbf{Semantic Priming} rather than compositional logic \citep{ettinger-2020-bert, kassner-schutze-2020-negated}. We hypothesize that the co-occurrence of an agent (e.g., \textit{carpenter}) and a telic verb (e.g., \textit{building}) activates the ``completed result'' representation in the model's high-dimensional space with such magnitude that it overrides the attention mechanism's focus on the subsequent cancellation clause. This phenomenon exemplifies a failure of contextual fidelity, where static parametric knowledge priors overpower the dynamic input context \citep{mallen-etal-2023-trust}.

\paragraph{The Outlier: \texttt{Qwen2.5} shows exceptional performance.}
\texttt{Qwen2.5} stands as a notable exception, maintaining a balanced performance with a TBR of $0.53$ and a healthy $\Delta_{AA}$ of $0.44$. Unlike its peers, Qwen demonstrates an emerging ability to suspend judgment in ambiguous contexts. 

\subsection{Trade-off: Bias Mitigation vs. Semantic Calibration in Advanced Prompts} 

Evaluating three advanced prompting strategies reveals a consistent trade-off between mitigating teleological bias and maintaining semantic calibration across aspectual classes.

\paragraph{DAP: The Knowledge-Application Gap.}
The DAP which introduces expert linguistic definition for the aspectual classes yields mixed results, exposing a critical gap between possessing linguistic knowledge and applying it. While supplying the linguistic rule improves \texttt{Llama-3.1}'s accuracy on Group C ($0.02 \to 0.36$), it remains ineffective for \texttt{Mistral} ($0.06$). 
This persistence of bias implies that the teleological prior is not a misunderstanding of the task definition but a robust feature of the model's probability distribution. As noted by \citet{dziri2024faith}, compositional reasoning in Transformers often degrades when it conflicts with high-frequency statistical patterns (i.e., the reporting bias) seen during pre-training.

\paragraph{CoT: Reasoning-Induced Uncertainty.}
CoT prompting successfully reduces the TBR for most models, proving that allocating compute to intermediate reasoning helps suppress System-1 heuristics \citep{wei2022chain}. 
However, this comes at a cost. We observe a significant degradation in Group D (Activities), where accuracy drops substantially (e.g., \texttt{Llama-3.1} $0.98 \to 0.65$). Qualitative analysis reveals a phenomenon of ``over-reasoning'' or unfaithful reasoning \citep{turpin-etal-2024-language}. When forced to analyze ``step-by-step'', models often hallucinate hypothetical interruptions for simple activities.\footnote{See, e.g., the activity ``\textit{running}'', with model reasoning: ``\textit{The premise describes a process in the present tense, indicating that the action of the athletes running on the track is ongoing or at least not explicitly completed.}''}  
CoT appears to over-generalize the progressive-as-incomplete heuristic: when forced to reason step-by-step, models often impose hypothetical interruptions even on atelic verbs, inadvertently suppressing valid entailments.

\begin{figure}[htbp]
    \centering
    \includegraphics[width=1\linewidth]{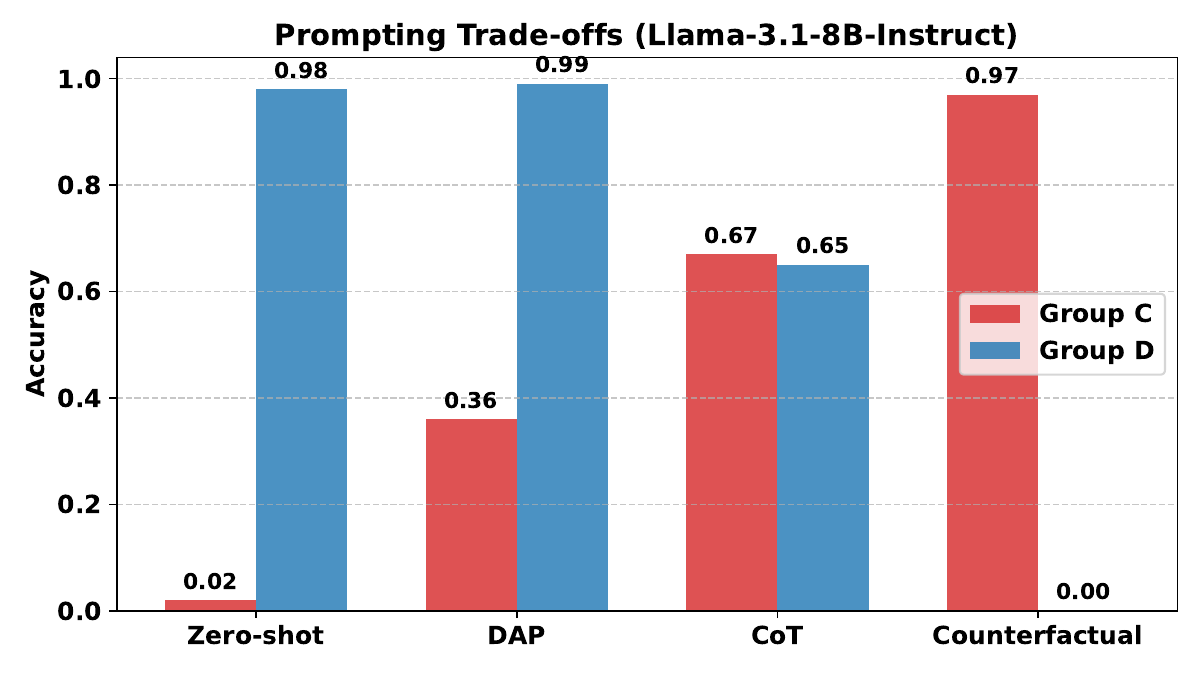} 
    \caption{The trade-off in prompting (\texttt{Llama-3.1}). As prompts become more aggressive in targeting bias (Zero-shot $\to$ Counterfactual), accuracy on Group C improves significantly (Pearson $r=1.00, p=0.00$),  while accuracy on Group D declines, though less consistently ($r=-0.91, p=0.09$).
    This trade-off illustrates the calibration bias.}
    \label{fig:tradeoff}
\end{figure}

\paragraph{Counterfactuals: The Calibration Crisis.}
The Counterfactual prompt represents the most aggressive intervention for arbitrary interruption. While highly effective at curing teleological bias (Group C Acc $\to 1.00$), it induces an obvious calibration bias. 
Accuracy on Group D collapses to near zero for \texttt{Llama-3.1}. We show the example of \texttt{Llama-3.1} in Figure \ref{fig:tradeoff}. 
The models swing from being ``naively optimistic'' (Zero-shot) to ``paranoidly skeptical'' (Counterfactual). The Counterfactual clearly over-instructed the models to follow the imperfective patterns of the telic verbs in Group C, resulting in misunderstanding for the atelic verbs in Group D. The models failed to dynamically adjust their reasoning strategy based on the (a)telicity of verbs, 
resulting in a wholesale rejection of progressive entailments. 
This highlights a lack of (aspectual) calibration \citep{kadavath2022language,NEURIPS2024_9c20f16b}: while the counterfactual prompt successfully suppresses teleological bias for telic verbs, models fail to modulate this skepticism based on verb class, applying it indiscriminately to atelic verbs as well. 
Detailed analyses and visualization for other models are provided in Appendix \ref{app:additional_results}.

\paragraph{Summary on Different Prompts.}
In summary, current LLMs struggle to balance the Imperfective Paradox. They oscillate between two extremes: \textbf{naive teleology} (Zero-shot) where everything is completed, and \textbf{paranoid skepticism} (Counterfactual) where nothing is ever certain (see the trends in Figure \ref{fig:tradeoff}). This inability to decouple \textit{telicity} from \textit{entailment} highlights a critical gap in the true aspectual reasoning capabilities of LLMs.

\subsection{Impact of Model Scale}
\label{sec:model_scale}

\begin{figure}[htbp]
    \centering
    \includegraphics[width=1\linewidth]{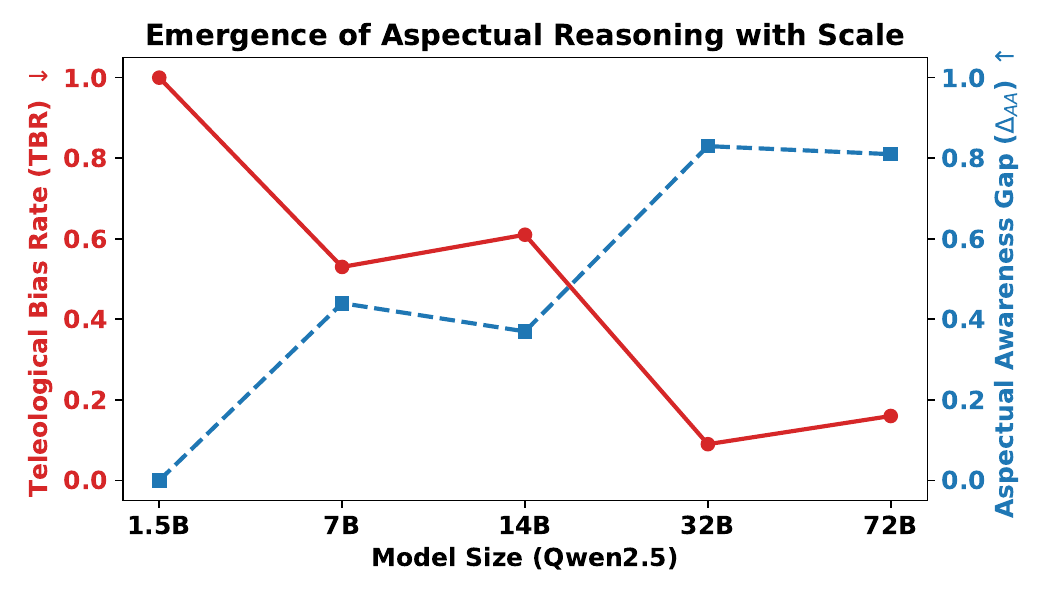}
    \caption{The emergence of aspectual reasoning capabilities across model scales. It shows a significant decrease in TBR and a corresponding increase in $\Delta_{AA}$.}
    \label{fig:size}
\end{figure}

To understand whether aspectual reasoning is a capability that improves linearly with scale or emerges as a distinct skill past a certain threshold, we evaluate the \texttt{Qwen2.5} family, which showed better performance in §\ref{sec:zero-shot}, across five distinct parameter sizes: 1.5B, 7B, 14B, 32B, and 72B. The results, detailed in Table \ref{tab:size} (Appendix) and visualized in Figure \ref{fig:size}, reveal that aspectual reasoning follows a distinct trajectory of emergence.

We observe a clear monotonic relationship between model scale and both evaluation metrics. Model size is strongly negatively correlated with Teleological Bias Rate (TBR; $r=-0.93, p=0.02$) and strongly positively correlated with the Aspectual Awareness Gap ($\Delta_{AA}$; $r=0.95, p=0.01$), indicating that increased model capacity systematically reduces teleological overgeneralization while enhancing aspect-sensitive reasoning.

The 1.5B model exhibits a complete inability to handle the Imperfective Paradox. With a TBR of $1.00$ and a $\Delta_{AA}$ of $0.00$, it treats every progressive verb as an entailed completion. 
As model scale increases, a dramatic ``phase shift''  \citep{wei2022emergent} is observed, especially at the 32B level:
Accuracy on Group C surges to $0.91$, and the $\Delta_{AA}$ peaks at $0.83$. This suggests that robust aspectual distinction, the ability to selectively inhibit the entailment for telic verbs while maintaining it for atelic ones, is an emergent ability. Although the 72B model maintains strong performance, $\Delta_{AA}$ decreases slightly to 0.81, suggesting a potential saturation or optimization bottleneck rather than further qualitative gains. Overall, these results indicate that aspectual understanding improves with scale and exhibits non-linear emergence characteristics rather than linear scaling.

\subsection{Fine-grained Analysis on Verb Semantics}
\label{sec:semantic_analysis}

\begin{figure}[htbp]
    \centering
    \includegraphics[width=1\linewidth]{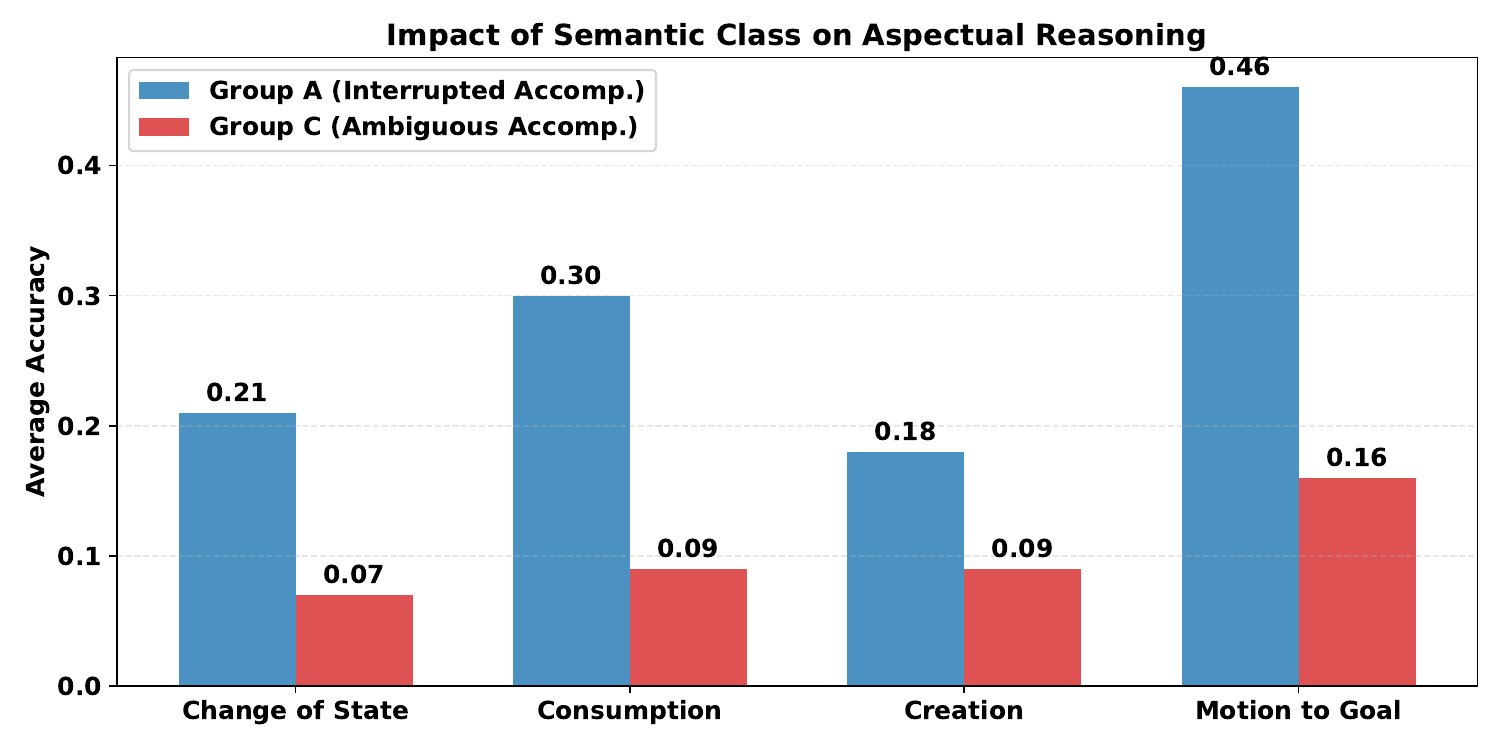}
    \caption{Average accuracy across models by semantic verb classes in Zero-shot. \textbf{Motion verbs} are consistently easier for models to reason about, while \textbf{Creation verbs} induce strong teleological bias (low accuracy in Group C) and resistance to contextual cancellation (lowest accuracy in Group A).}
    \label{fig:verb_analysis}
\end{figure}

Do all telic verb classes 
trigger the same degree of teleological bias? To answer this, we disaggregate the performance on Group A and Group C based on the four semantic sub-classes defined in \S\ref{sec:dataset}: \textit{Creation}, \textit{Consumption}, \textit{Change of State}, and \textit{Motion to Goal}. Figure \ref{fig:verb_analysis} visualizes the average accuracy across all evaluated models in the zero-shot setting. Detailed results for each model as well as for other prompt settings can be found in Appendix \ref{app:additional_results}.

\paragraph{The Resilience of Motion vs. The Creation Penalty.}
We observe a significant semantic divergence. Verbs of \textit{Motion to Goal} (e.g., ``\textit{cross}'') consistently yield the highest performance. In Group A, the average accuracy for \textit{Motion to Goal} verbs is $\mathbf{46\%}$, significantly outperforming \textit{Creation} ($\mathbf{18\%}$) and \textit{Change of State} ($\mathbf{21\%}$). The performance gap is substantial: $\Delta_{Motion-Creation} \approx +28\%$ in Group A and $\approx +7\%$ in Group C.\footnote{We additionally conducted a two-way repeated-measures ANOVA for Group~A and Group~C across models and semantic verb classes. The results revealed a significant main effect of verb class ($F(3,18)=6.98, p<0.01$). Crucially, we observed a significant interaction between group and verb class ($F(3,18)=3.88, p<0.05$), indicating that the impact of ambiguity varies across semantic classes.}

We hypothesize that this stems from the nature of the resultative state. For \textit{Creation} verbs (e.g., ``\textit{build}''), the result involves the existential realization of an object. This explicit ``coming into being'' appears to act as a stronger attractor in the language model's predictive distribution than a mere change in spatial coordinates involved in \textit{Motion}. Consequently, models exhibit higher resistance to negation when an artifact's existence is at stake.

\paragraph{Representational Analysis: Encoding vs. Inference.}
Does teleological bias reflect a failure to distinguish \textit{process} from \textit{result} in the model's internal representations, or does it arise later, at the point of inference? In order to probe this, we measured the cosine similarity between contextual embeddings of progressive (e.g., \textit{``was building''}) and perfective (e.g., \textit{``built''}) verb phrases using \texttt{bert-base-uncased} \citep{devlin-etal-2019-bert}, 
mean-pooled over the word pieces of each verb phrase, treating similarity as a proxy for how well an encoder separates the two aspectual forms.

\begin{figure}[htbp]
    \centering
    \subfloat[Cosine similarity  \\\ (Process vs. Result).]{
        \includegraphics[width=0.49\linewidth]{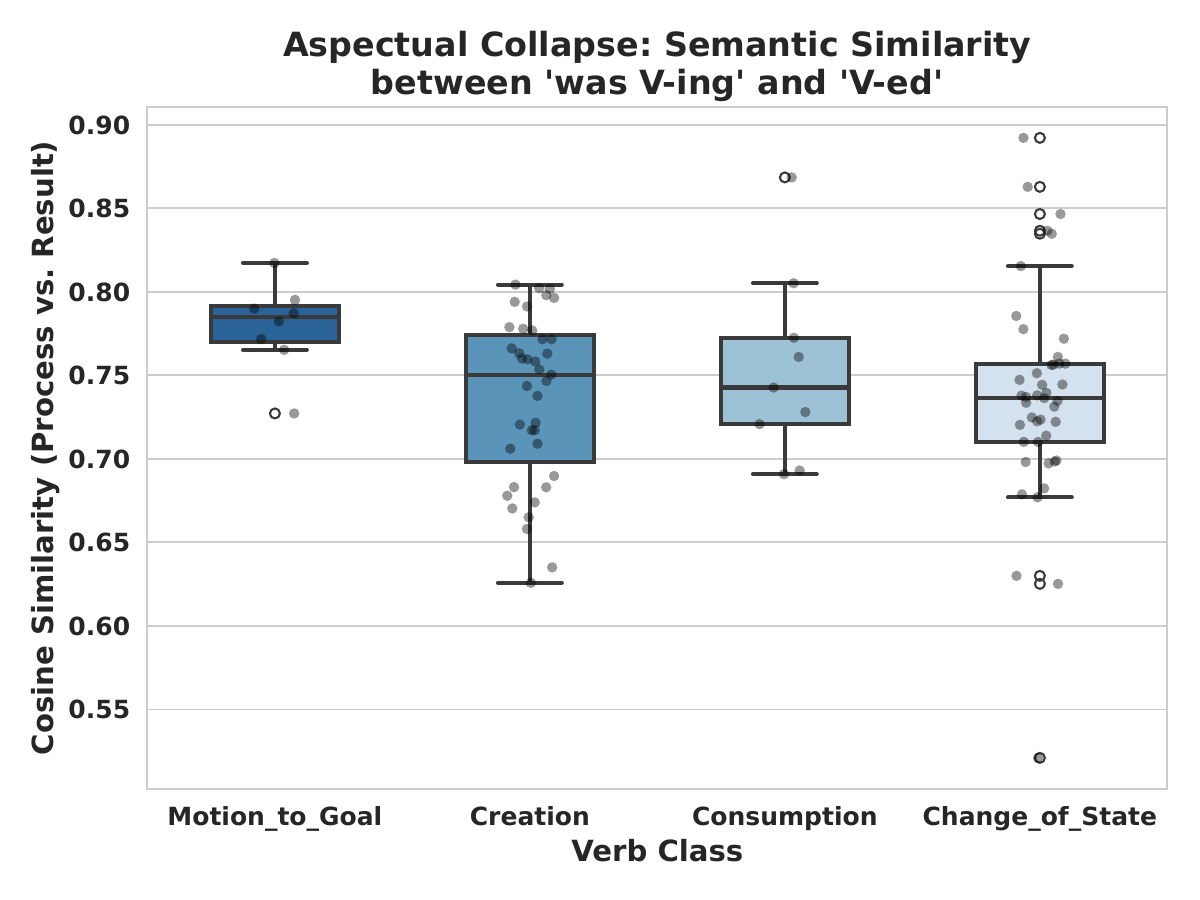}
        \label{fig:rep_boxplot}
    }
    \subfloat[\small{Similarity vs. TBR$_C$ \\\ (Zero-shot).}]{
        \includegraphics[width=0.49\linewidth]{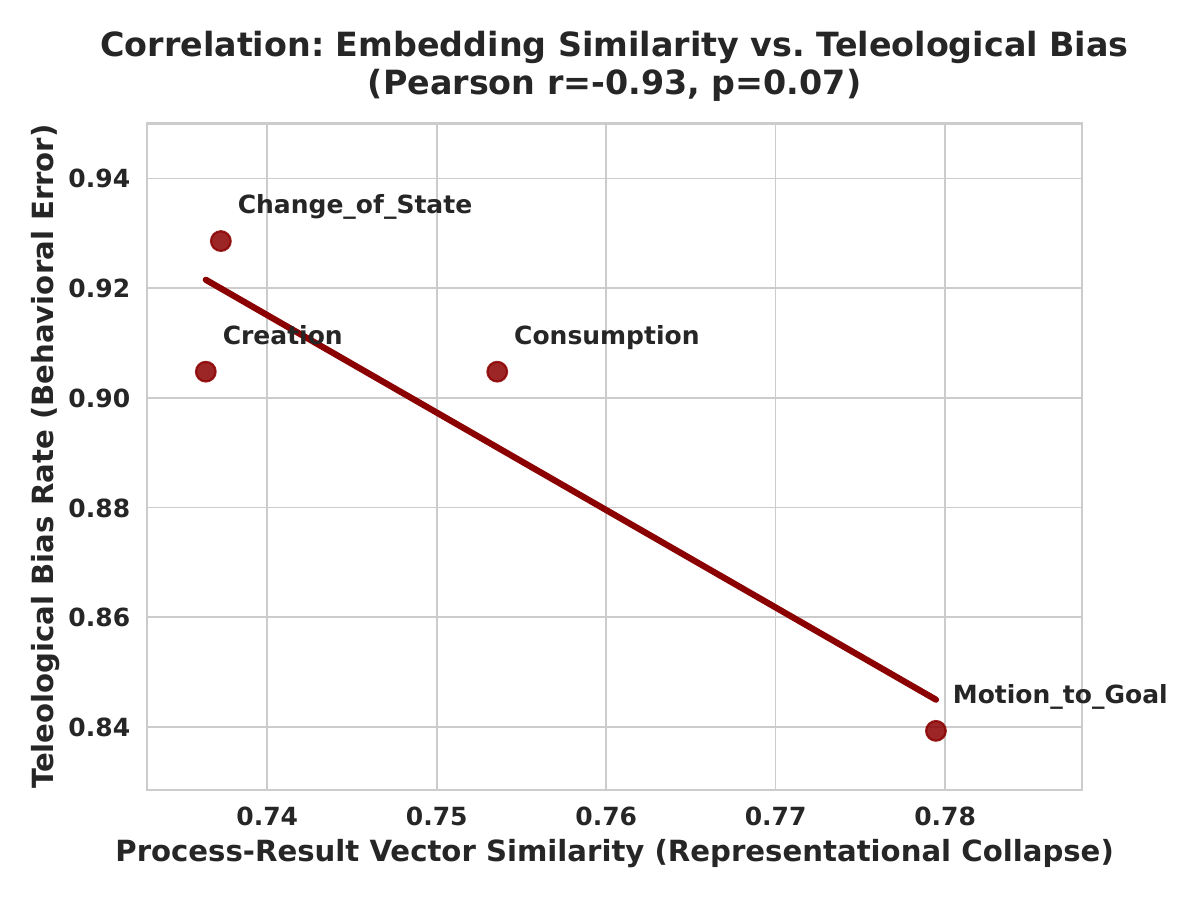}
        \label{fig:rep_correlation}
    }
    \caption{Representational vs. Behavioral Divergence.
    }
    \label{fig:representational_analysis}
\end{figure}

Figure \ref{fig:representational_analysis} reveals an inverse relationship (Pearson $r=-0.93, p=0.07$) between how similarly the progressive and perfective forms of a verb are encoded, and how strongly models hallucinate completion for that verb class. Crucially, this relationship runs in the opposite direction from what a simple confusion account would predict. \textit{Motion to Goal} verbs show relatively higher similarity (median $\approx$ 0.78), meaning ``\textit{was crossing}'' and ``\textit{crossed}'' are encoded similarly; yet models reason about them most accurately (with TBR$_C$ being the lowest). \textit{Creation} verbs show only slightly lower similarity (median $\approx$ 0.75), meaning the encoder does distinguish ``\textit{was building}'' from ``\textit{built}''; yet these verbs trigger higher hallucination rates (with TBR$_C$ being higher).

This dissociation suggests that the bias does not stem from the model failing to encode the aspectual distinction. If it did, we would expect high similarity to correlate with high error, but we observe the opposite trend. Instead, even when the representation separates process from result, the inference step is overridden by a strong prior that \textit{Creation} events reach their goal. The difference between ``\textit{was building}'' and ``\textit{built}'' is encoded, but models still predict completion, because their training has instilled a robust expectation that construction projects succeed. We therefore hypothesize that Teleological Bias is not a failure of \textit{perception} (encoding), but a failure of \textit{reasoning} (decoding): the model's decoding process is captured by world-knowledge priors rather than guided by the logical content of the input.

\section{Conclusion}
In this work, we evaluated and discovered the \textbf{Teleological Bias} in LLMs using our diagnostic dataset, \textsc{ImperfectiveNLI}. Our findings indicate that 
current models fundamentally struggle to resolve the \textbf{Imperfective Paradox}, frequently conflating the \textit{process} of a telic event with its \textit{culmination}. 
Our investigation into prompting strategies reveals a critical calibration crisis, finding that while some strategies reduce overconfident or hallucinated predictions, they can also make models overly cautious, causing them to reject correct entailments.
Meanwhile, this aspectual understanding can improve with model scale     unevenly.  
Furthermore, our analysis suggests that this bias is not merely a representational deficit but a decision-time failure: while internal embeddings often distinguish between process and result, strong probabilistic priors regarding goal attainment override these signals during inference. 
We conclude that solving aspectual reasoning requires moving beyond inference-time prompting toward aligning models to structurally respect the internal logical boundaries of event semantics. 
We hope that \textsc{ImperfectiveNLI} provides a foundation for future work on targeted alignment training, cross-lingual aspectual reasoning, and the broader challenge of teaching LLMs to respect formal semantic boundaries.

\section{Limitations}

While our work offers important insights into the imperfective paradox with the  teleological bias of LLMs, we acknowledge several limitations that define the scope of our findings and point towards future research directions.

\paragraph{Ecological Validity and Syntactic Diversity.}
To strictly isolate the variable of aspectual telicity, we employed a template-based approach to construct \textsc{ImperfectiveNLI}. While this ensures high internal validity and precise minimal-pair comparisons, it sacrifices lexical and syntactic diversity. Real-world language usage often includes complex discourse markers, temporal adjuncts (e.g., \textit{``for an hour''} vs. \textit{``in an hour''}), and pragmatic cues that constrain interpretation to express aspectuality \cite{zhang1995contrastive}. 
Consequently, our controlled setup likely overestimates model competence relative to naturalistic settings; handling the imperfective paradox in messier, more complex narratives may prove considerably more challenging.

\paragraph{Linguistic Specificity.}
Our study focuses exclusively on English, a language that marks the progressive aspect periphrastically (via auxiliary ``\textit{be}'' + ``\textit{-ing}''). However, aspect is a typologically diverse category. For instance, Slavic languages mark aspect morphologically (aspect prefixes), while Mandarin Chinese uses aspectual particles. 
Since LLM tokenizers and representations vary across languages, it remains an open question whether the teleological bias is a universal cognitive artifact of Transformers or an artifact of English-centric pre-training data. Meanwhile, in human cognitive studies, research demonstrates that the conceptualization of event semantics differs fundamentally between aspect languages (those with grammaticalized aspect markers, e.g., English, Chinese, Russian) and non-aspect languages (e.g., German) \cite{vonStutterheim2012,FLECKEN2014}. Consequently, extending this investigation to a cross-lingual and typological framework represents a vital direction for future research.

\paragraph{Theoretical Linguistics vs. Human Judgments.}
Since our paper focuses on the theoretical concepts of imperfective paradox, we note potential divergence between theoretical labels and empirical human judgments. As the first study systematically investigating the Imperfective Paradox in LLMs, our primary objective was to characterize the teleological bias, the models' tendency to hallucinate completion, based on established theoretical semantics. While we focus here on establishing this bias in models, we wish to note that exploring the fine-grained nuances of human perception in these contexts is a vital next step.

\paragraph{Inference-Only Interventions.}
Our mitigation strategies were limited to inference-time prompting (CoT, Counterfactuals). While we identified the performance trade-offs as a side effect of these prompts, we did not explore parameter-efficient fine-tuning (PEFT) or activation steering techniques to permanently alter the model's internal aspectual representations. Future work should investigate whether the ``dissociation'' between representation and reasoning can be bridged through targeted alignment training rather than transient prompting. 

\paragraph{Scope of World Knowledge.}
Finally, our semantic analysis focused on four broad classes of telic verbs. We did not model the granularity of ``script knowledge'' \citep{schank1977scripts,lyu-etal-2021-goal}. For example, the probability of interruption varies by event type (e.g., ``\textit{building a house}'' takes months and is prone to interruption, whereas ``\textit{drawing a circle}'' takes seconds). Current models may rely on these latent temporal priors, which our uniform evaluation setup does not explicitly disentangle from grammatical aspect.

\section{Ethical Considerations}

\paragraph{Human Subjects for Data Evaluation.} 
For the human evaluation of sentence quality, we recruited external participants via the \emph{Prolific} platform. Prior to the task, annotators were fully informed about the study's scope and provided informed consent. We ensured that no personally identifiable information (PII) was collected. Evaluators were compensated at an hourly rate of £9, exceeding the platform's recommended minimum and comparable to the local living wage. 

\paragraph{Artifact Usage.} 
Regarding the curated artifacts, we release the \textsc{ImperfectiveNLI} dataset under the Creative Commons Attribution-NonCommercial 4.0 International License (CC BY-NC 4.0). 
This license explicitly permits the distribution and use of the dataset for research purposes. 
We confirm that our use of these existing artifacts is consistent with their intended academic use, and our derivative dataset is released with the strict specification that it be used solely within research contexts, ensuring compatibility with the original access conditions. The dataset contains no harmful or offensive content.

\paragraph{Use of AI Assistants.} 
As described in Section \ref{sec:dataset}, we used Gemini to assist in the initial 
generation and curation of the dataset. To ensure data quality and mitigate potential machine artifacts, all generated instances underwent rigorous manual verification and refinement by three native English speakers (see validation details in Section \ref{sec:dataset} and Appendix \ref{app:human}). 

Regarding the manuscript and experimental code, we acknowledge the use of ChatGPT solely for grammatical correction, stylistic polishing, and assistance with preliminary coding routines. The authors verified all AI-assisted outputs and bear full responsibility for the accuracy and originality of the content presented in this work.

\section*{Acknowledgments}

This publication was supported by LMUexcellent, funded by the Federal Ministry of Research, Technology and Space (BMFTR) and the Free State of Bavaria under the Excellence Strategy of the German Federal Government and the Länder.

We also thank Max Müller-Eberstein for valuable discussions and insightful exchanges related to this project.



\bibliography{custom}


\appendix

\section{Verbs}

We show here the telic verbs and atelic verbs probed in the \textsc{ImperfectiveNLI} and in the main experiment:

\textbf{Atelic verbs:} ['run', 'swim', 'walk', 'push', 'chat', 'dance', 'sing', 'laugh', 'weep', 'shout', 'play', 'wander', 'discuss', 'wait', 'study', 'work', 'exercise', 'browse', 'float', 'glide', 'slide', 'roll', 'snore', 'dream', 'meditate', 'tremble', 'sweat', 'bleed', 'shine', 'vibrate', 'hum', 'roar', 'bark', 'stare', 'listen', 'watch', 'cuddle', 'fight', 'argue', 'wrestle', 'garden', 'shop', 'patrol', 'chase', 'jog', 'march', 'ski', 'surf', 'ride', 'whistle', 'smile', 'frown', 'scream', 'yell', 'whisper', 'mumble', 'talk', 'speak', 'gossip', 'tease', 'help', 'serve', 'rule', 'live', 'breathe', 'cough', 'yawn', 'blink', 'nod', 'clap', 'shiver', 'twitch', 'ache', 'itch', 'glow', 'sparkle', 'fade', 'rot', 'bloom', 'grow', 'shrink', 'spin', 'rotate', 'travel', 'commute', 'roam', 'drift', 'flow', 'rain', 'snow', 'blow', 'graze', 'hunt', 'fish', 'cook', 'smoke', 'sip', 'chew', 'rub', 'scratch']

\textbf{Telic verbs: }

Change of State: ['fix', 'fill', 'clean', 'repair', 'empty', 'dry', 'wash', 'iron', 'sharpen', 'inflate', 'freeze', 'melt', 'cure', 'renovate', 'demolish', 'dig', 'pack', 'unlock', 'memorize', 'learn', 'organize', 'erase', 'delete', 'install', 'wrap', 'open', 'close', 'rescue', 'save', 'mow', 'sweep', 'mop', 'polish', 'sand', 'crack', 'cut', 'slice', 'chop', 'boil', 'fry', 'roast', 'toast', 'decorate', 'furnish']

Consumption: ['eat', 'drink', 'read', 'burn', 'devour', 'spend', 'destroy', 'solve', 'peel']

Creation: 
['build', 'paint', 'bake', 'write', 'knit', 'compose', 'draw', 'film', 'code', 'sew', 'sculpt', 'weave', 'brew', 'forge', 'design', 'prove', 'translate', 'construct', 'invent', 'produce', 'plan', 'calculate', 'measure', 'count', 'transcribe', 'download', 'upload', 'print', 'plant', 'harvest', 'mix', 'equip', 'establish', 'form', 'shape', 'assemble', 'gather', 'collect', 'pick']

Motion to Goal:
['climb', 'cross', 'hike', 'fly', 'mail', 'deliver', 'catch', 'locate']

\section{Prompts}
\label{app:prompt}

In this section, we present the detailed prompt templates used across our four experimental conditions. The placeholders \texttt{\{premise\}} and \texttt{\{hypothesis\}} are dynamically replaced with the corresponding instances from the dataset. The specific instructions and output formatting constraints for each strategy are shown below.

\begin{tcolorbox}[colback=gray!5!white, colframe=gray!75!black, title=\small \textbf{P1: Strict Logic Prompt}, boxrule=0.8pt, left=2pt, right=2pt, top=2pt, bottom=2pt]
\footnotesize
\textbf{System Prompt:} You are a strict logician. Your task is to determine if a Hypothesis is necessarily true given a Premise.
\begin{itemize}[noitemsep,topsep=0pt,leftmargin=*]
    \item If the Hypothesis MUST be true based only on the Premise, output \textbf{"True"}.
    \item If the Hypothesis is contradicted by the Premise, output \textbf{"False"}.
    \item If the Hypothesis might be true but is not explicitly guaranteed by the Premise, output \textbf{"Unknown"}.
\end{itemize}
Do not use common sense assumptions. Only use the text provided.

\textbf{User Input:} Premise: \{premise\} Hypothesis: \{hypothesis\} \\
Please respond with ONLY one of the following options: "True", "False", or "Unknown".
\end{tcolorbox}

\begin{tcolorbox}[colback=gray!5!white, colframe=gray!75!black, title=\small \textbf{P2: Definition-Aware Prompt}, boxrule=0.8pt, left=2pt, right=2pt, top=2pt, bottom=2pt]
\footnotesize
\textbf{System Prompt:} ... [Standard Logic Instructions] ...

\textbf{Important Linguistic Rule:} Note the distinction between activity and accomplishment verbs.
\begin{itemize}[noitemsep,topsep=0pt,leftmargin=*]
    \item For \textbf{accomplishments/goal-oriented actions} (e.g., "was building", "was writing"), the progressive form does NOT imply completion.
    \item For \textbf{activities} (e.g., "was running", "was singing"), the process implies the action occurred.
\end{itemize}

\textbf{User Input:} Premise: \{premise\} Hypothesis: \{hypothesis\} \\
Please respond with ONLY one of the following options: "True", "False", or "Unknown".
\end{tcolorbox}

\begin{tcolorbox}[colback=gray!5!white, colframe=gray!75!black, title=\small \textbf{P3: Chain-of-Thought Prompt}, boxrule=0.8pt, left=2pt, right=2pt, top=2pt, bottom=2pt]
\footnotesize
\textbf{System Prompt:} [Same as Strict Logic Prompt]

\textbf{User Input:} Premise: \{premise\} Hypothesis: \{hypothesis\} \\
\textbf{Instruction:} First, analyze the temporal status of the event in the premise. Does the action have a defined endpoint? Was it completed? Then, provide your final label.

\textbf{Output Requirement:} You must output your response in strict \textbf{JSON format} containing exactly two keys:
\begin{itemize}[noitemsep,topsep=0pt,leftmargin=*]
    \item \texttt{"reasoning"}: A string explaining your step-by-step logic.
    \item \texttt{"label"}: The final judgment ("True", "False", or "Unknown").
\end{itemize}
Do not include markdown formatting (like ```json). Return ONLY the JSON object.

\textbf{Example Format:} \\
\{"reasoning": "The premise describes a process...", \\
  "label": "True" / "False" / "Unknown" 
\}
\end{tcolorbox}

\begin{tcolorbox}[colback=gray!5!white, colframe=gray!75!black, title=\small \textbf{P4: Counterfactual Simulation Prompt}, boxrule=0.8pt, left=2pt, right=2pt, top=2pt, bottom=2pt]
\footnotesize
\textbf{System Prompt:} [Same as Strict Logic Prompt]

\textbf{User Input:} Premise: \{premise\} Hypothesis: \{hypothesis\} \\
\textbf{Instruction:} 
1. First, list 3 possible real-world scenarios where the action in the premise occurs but the result in the hypothesis is NOT achieved (e.g., interruptions, failures).
2. Based on these possibilities, determine if the hypothesis is \textit{necessarily} true.

\textbf{Output Requirement:} You must output your response in strict \textbf{JSON format} containing exactly two keys:
\begin{itemize}[noitemsep,topsep=0pt,leftmargin=*]
    \item \texttt{"possible\_interruptions"}: a list of three strings for 3 possible real-world scenarios.
    \item \texttt{"label"}: The final judgment ("True", "False", or "Unknown").
\end{itemize}
Do not include markdown formatting (like ```json). Return ONLY the JSON object.

\textbf{Example Format:} \\
\{ 
  "possible\_interruptions": ["..."], \\
  "label": "True" / "False" / "Unknown" 
\}
\end{tcolorbox}

\section{Human Evaluation and Results}
\label{app:human}

To validate the quality of the constructed dataset, we conducted a human evaluation on the \textit{Prolific} crowdsourcing platform. We recruited 3 native English speakers to evaluate the generated sentences.

Following established protocols in machine translation and text generation research \cite{chen-etal-2022-mtg, wu-etal-2025-absa}, evaluators were instructed to rate each sentence on a 5-point Likert scale across three linguistic dimensions: \textit{Grammar}, \textit{Fluency}, and \textit{Adequacy}. To ensure high-quality feedback, evaluators were also encouraged to provide corrections for any sentence rated 3 or lower. The specific instructions provided to the annotators are detailed in the following box: 

\begin{tcolorbox}[
    colback=gray!5!white, 
    colframe=blue!55!black, 
    title=\small \textbf{Instructions Provided to Prolific Evaluators}, 
    boxrule=0.8pt, 
    left=4pt, right=4pt, top=4pt, bottom=4pt,
    label={box:instructions} 
]
\footnotesize
\textbf{Task Overview:} Please evaluate the English sentences provided in the attached Excel file. For each sentence, select a score from 1 to 5 using the dropdown menu. Please do not copy-paste values.

\vspace{0.5em}
\textbf{Rating Criteria:}
\begin{itemize}[noitemsep,topsep=0pt,leftmargin=*]
    \item \textbf{Grammar}: Is the sentence grammatically correct? \\
    (5 = Perfect, no errors; 1 = Major errors, grammatically broken)
    
    \item \textbf{Fluency}: Does the sentence sound natural to a native speaker? \\
    (5 = Very natural; 1 = Very awkward/unnatural)
    
    \item \textbf{Adequacy}: Is the meaning clear and logical? \\
    (5 = Perfectly clear; 1 = Confusing or nonsensical)
\end{itemize}

\vspace{0.5em}
\textbf{Correction (Optional)}: If you assign a score of 3 or lower in any category, please rewrite the sentence in the "Proposed Correction" column to improve its quality.
\end{tcolorbox}

\begin{table}[htbp]
\centering
\small
\renewcommand{\arraystretch}{1.2}
\begin{tabular}{lccc}
\toprule
\textbf{Metric} & \textbf{Mean Score} & \textbf{Std. Dev.} & \textbf{Agreement (\%)} \\ \midrule
Grammar  & 4.63 & 0.49 & 99.8\% \\
Fluency  & 4.53 & 0.56 & 97.7\% \\
Adequacy & 4.46 & 0.58 & 97.2\% \\ \midrule
\textbf{Average} & \textbf{4.54} & \textbf{-} & \textbf{98.2\%} \\ \bottomrule
\end{tabular}
\caption{Human evaluation results.  
``Agreement'' denotes the percentage of instances where the scores from two evaluators differed by at most 1 point (on a 5-point scale), averaged over all evaluator pairs.}
\label{tab:human_eval}
\end{table}

The results of the human evaluation are summarized in Table~\ref{tab:human_eval}. 
The generated dataset demonstrated exceptional quality across all linguistic dimensions. Specifically, the \textit{Grammar} score reached a mean of \textbf{4.63}, while \textit{Fluency} and \textit{Adequacy} achieved \textbf{4.53} and \textbf{4.46}, respectively. 
It is worth noting that due to the high concentration of scores at the top of the scale (ceiling effect), standard correlation metrics (e.g., Pearson's $r$ or Fleiss' $\kappa$) are not applicable as they require variance to be meaningful \cite{Feinstein1990,Gwet2008}. 
Consequently, instead of relying on variance-dependent metrics, we report the inter-annotator agreement rate, specifically calculating the percentage of sample pairs where the score difference between evaluators is within a tolerance of 1 (i.e., \textit{Percent Adjacent Agreement}). This approach provides a more robust measure of consensus for skewed data \cite{Stemler2004IRR,Williamson2012}.
The high agreement rates (averaging \textbf{98.2\%}) confirm that the evaluators reached a strong consensus on the high quality of the data.

In addition, for the few low-scoring or unnatural sentences, evaluators provided corrections which were adopted after manual verification.

\section{Implementation Details}
\label{sec:Implementation Details}
All experiments were run on a single NVIDIA A100 (80GB) GPU. The models are open-weight and directly downloaded from the Huggingface \texttt{Transformers} \cite{wolf-etal-2020-transformers}. 
The models are used solely for research purposes in accordance with their original terms, consistent with their intended use. And we do not redistribute any third-party model weights. 
We disable sampling and use greedy decoding to ensure deterministic outputs for all open LLMs by setting \texttt{do\_sample=False}. 
All generations are capped at 512 tokens.

\section{Additional Results}
\label{app:additional_results}

\paragraph{\textit{Qwen2.5} with Different Scales.} Table \ref{tab:size} presents detailed results of \textit{Qwen2.5} with different scales in Zero-shot. A sharp phase transition is observed between 14B and 32B, where the Aspectual Awareness Gap ($\Delta_{AA}$) surges by 124\% ($0.37 \to 0.83$), significantly outperforming the linear scaling trend ($r=0.79$).

\begin{table}[htbp]
\centering
\setlength\tabcolsep{3pt}
\renewcommand\arraystretch{0.9}
\small
\begin{tabular}{ccccccc}
\toprule
\textbf{Size} & \textbf{Acc A} & \textbf{Acc B} & \textbf{Acc C} & \textbf{Acc D} & \textbf{TBR$_C$} ($\downarrow$) & \textbf{$\Delta_{AA}$} ($\uparrow$) \\
\midrule
1.5B & 0.21 & 0.96 & 0.00 & 1.00 & 1.00 & 0.00 \\
7B & 0.20 & 0.98 & 0.47 & 0.97 & 0.53 & 0.44 \\
14B & 0.24 & 0.86 & 0.39 & 0.98 & 0.61 & 0.37 \\
32B & 0.53 & 0.90 & 0.91 & 0.92 & 0.09 & 0.83 \\
72B & 0.43 & 0.88 & 0.84 & 0.97 & 0.16 & 0.81 \\
\bottomrule
\end{tabular}
\caption{Performance of the \textit{Qwen2.5} family across different parameter scales (Zero-shot). }
\label{tab:size}
\end{table}

\begin{figure}[!htbp]
    \centering
    \includegraphics[width=1\linewidth]{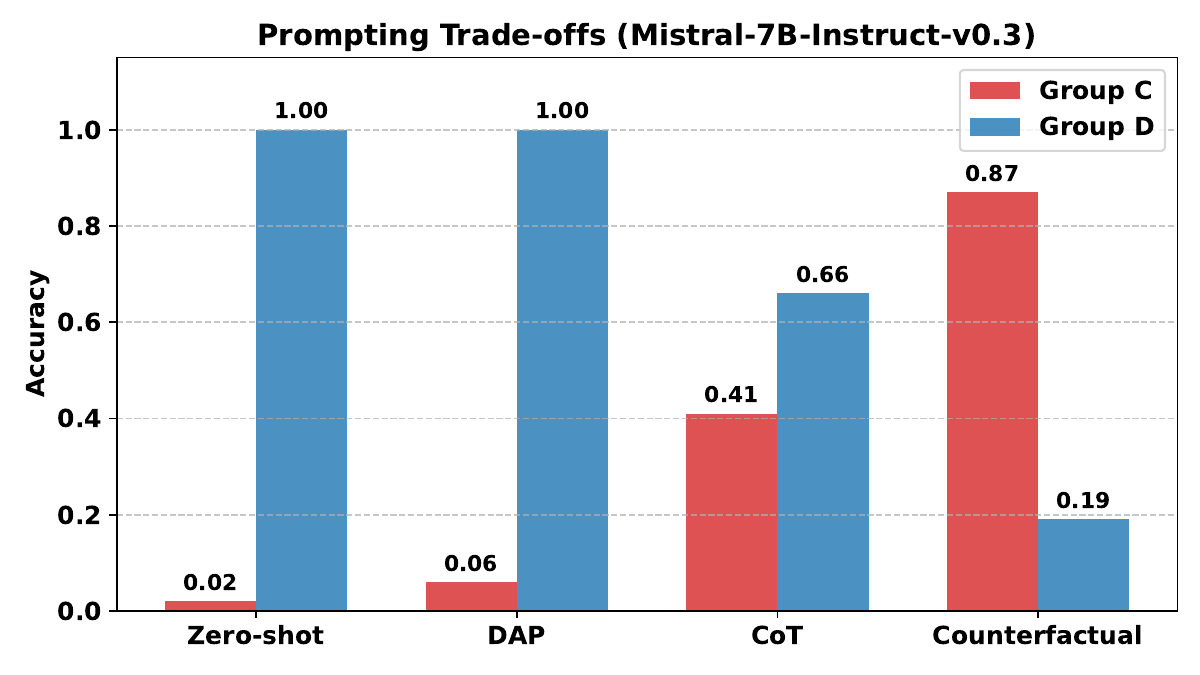}
    \caption{The Trade-off in Prompting (\texttt{Mistral}). The model shows a clear skepticism trap with strong correlations. (Group C: $r=0.95, p=0.05$; Group D: $r=-0.93, p=0.07$).} 
    \label{fig:tradeoff_mistral}
\end{figure}

\begin{figure}[!htbp]
    \centering
    \includegraphics[width=1\linewidth]{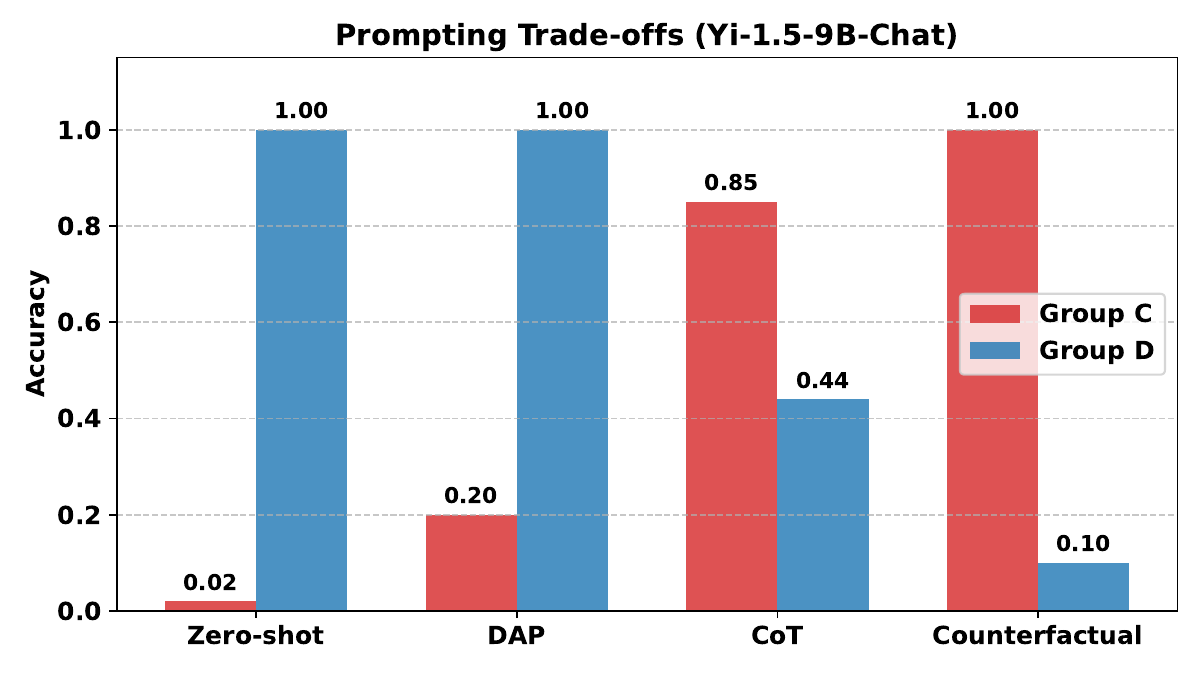}
    \caption{The Trade-off in Prompting (\texttt{Yi-1.5}). Similar to Llama, Yi exhibits a sharp trade-off. (Group C: $r=0.97, p=0.03$; Group D: $r=-0.95, p=0.05$).} 
    \label{fig:tradeoff_yi}
\end{figure}

\begin{figure}[!htbp]
    \centering
    \includegraphics[width=1\linewidth]{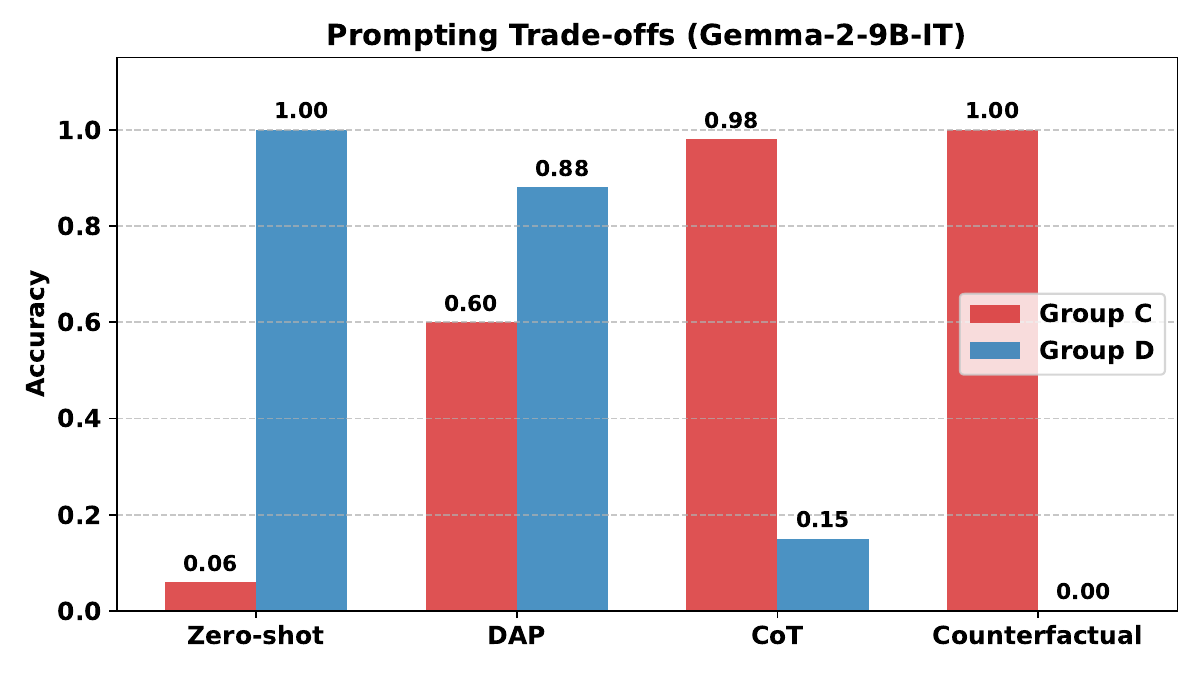}
    \caption{The Trade-off in Prompting (\texttt{Gemma-2}). The model achieves perfect correction on Group C but collapses to zero on Group D. (Group C: $r=0.94, p=0.06$; Group D: $r=-0.95, p=0.05$).} 
    \label{fig:tradeoff_gemma}
\end{figure}

\begin{figure}[!htbp]
    \centering
    \includegraphics[width=1\linewidth]{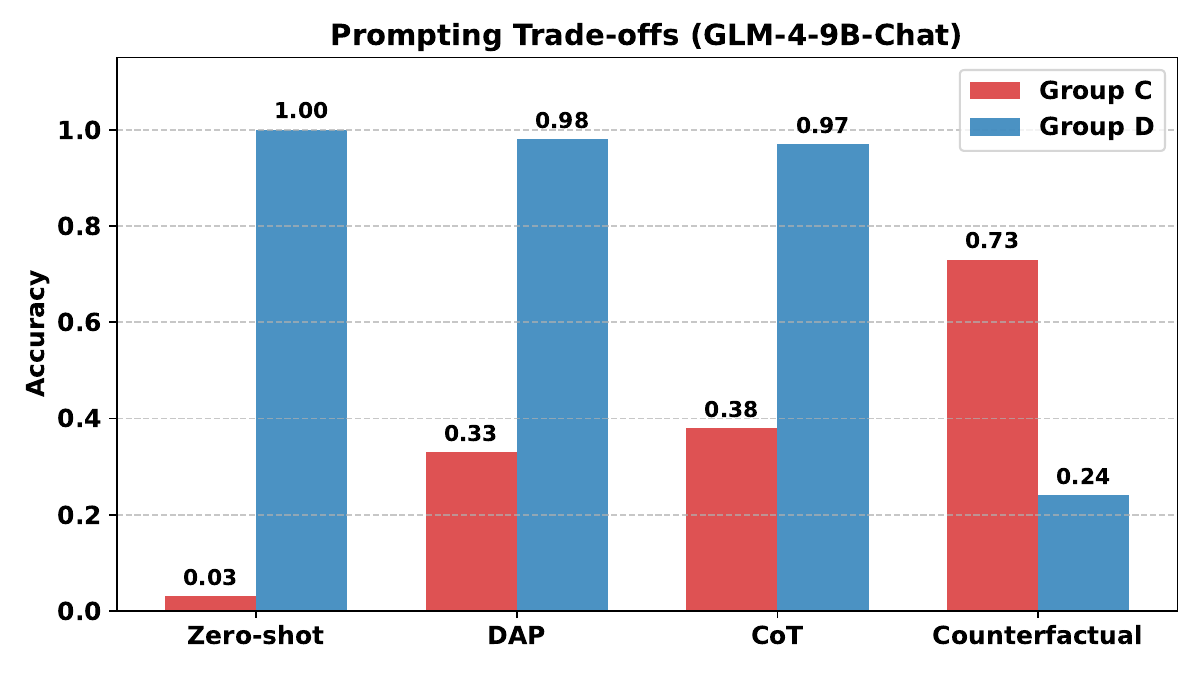}
    \caption{The Trade-off in Prompting (\texttt{GLM-4}). (Group C: $r=0.97, p=0.03$; Group D: $r=-0.79, p=0.21$).} 
    \label{fig:tradeoff_glm}
\end{figure}

\begin{figure}[!htbp]
    \centering
    \includegraphics[width=1\linewidth]{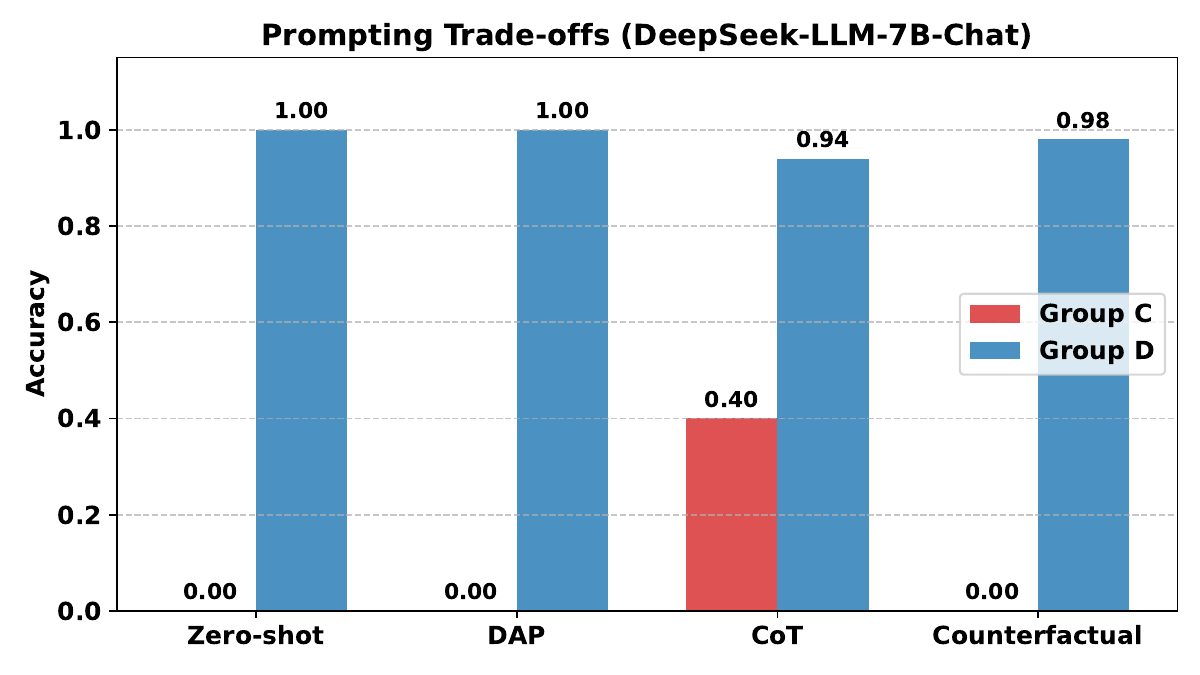}
    \caption{Resistance to Prompting (\texttt{DeepSeek}). Unlike other models, DeepSeek shows weak correlation, indicating rigidity against the counterfactual prompts. (Group C: $r=0.26, p=0.74$; Group D: $r=-0.55, p=0.45$).} 
    \label{fig:tradeoff_deepseek}
\end{figure}

\begin{figure}[!htbp]
    \centering
    \includegraphics[width=1\linewidth]{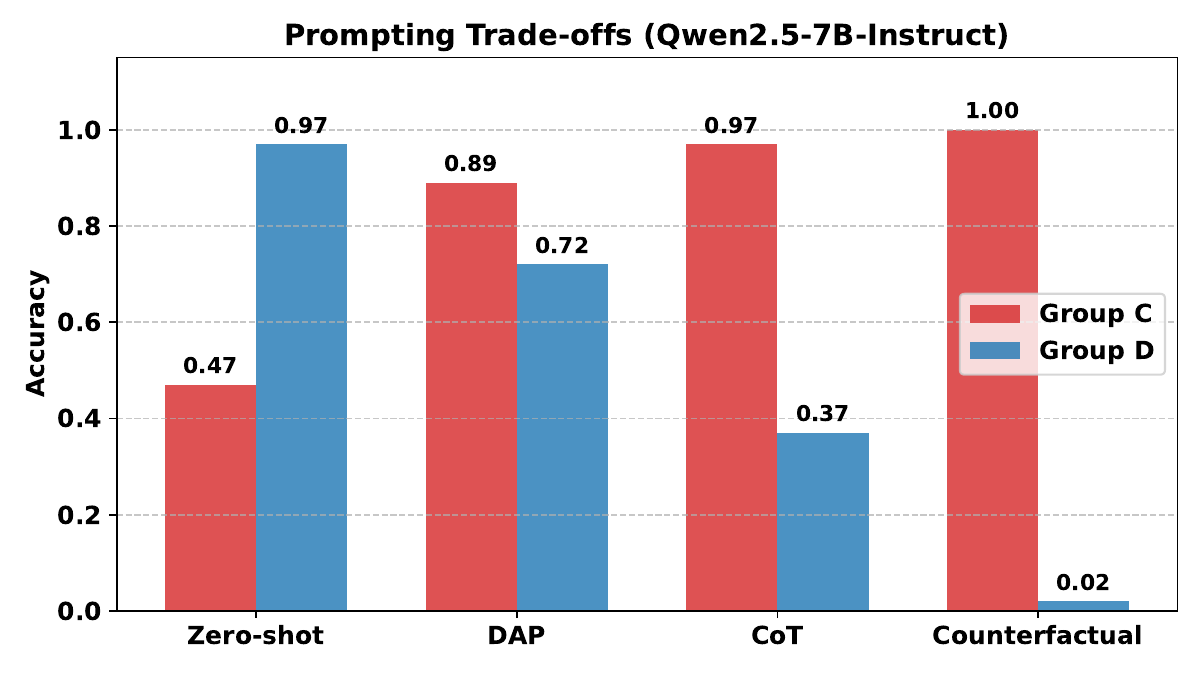}
    \caption{The Trade-off in Prompting (\texttt{Qwen2.5}). (Group C: $r=0.88, p=0.12$; Group D: $r=-1.00, p=0.00$).} 
    \label{fig:tradeoff_qwen}
\end{figure}

\paragraph{Trade-offs of the Prompts across Models.}
In this section, we extend the analysis of the trade-offs in prompting to the remaining six models. As illustrated in Figures \ref{fig:tradeoff_mistral}--\ref{fig:tradeoff_qwen}, the trade-off 
is pervasive. Models such as \texttt{Mistral-7B}, \texttt{Yi-1.5}, \texttt{Gemma-2}, and \texttt{GLM-4} exhibit inverse correlations. As the prompting strategy becomes more aggressive (Zero-shot $\to$ Counterfactual), these models successfully correct the teleological bias in Group C (Pearson $r > 0.90$) but suffer a catastrophic drop in performance on the atelic baseline in Group D (Pearson $r \leq -0.79$). This confirms that the calibration crisis is a general issue in current LLMs: they struggle to isolate the intervention to only the target telic verbs. \texttt{DeepSeek-LLM} (Figure \ref{fig:tradeoff_deepseek}) presents an outlier behavior. It shows resistance to the intervention, with no significant improvement on Group C ($p=0.74$) and consequently less degradation on Group D compared to others. This suggests a form of ``reasoning rigidity'' rather than the ``over-correction'' seen in other models.

\paragraph{The Predictions of “False” in Group C.}
Table \ref{tab:false_rate_group_c} additionally reports the rate of ``False'' predictions across all experimental settings. It turns out that very few models in very few cases predicted ``False''. This is reasonable as in such cases of Imperfective Paradox, the models would either select ``Unknown'' (the correct predictions for uncertainty) or ``True'' (the teleological bias). If models predicted ``False'', it could be interpreted as indicating that the models ``saw'' the past progressive for accomplishments as non-completion, which suggests over-prediction. However, based on our observation with the following table, this is not the case.

\begin{table}[t]
\centering
\small
\begin{tabular}{l l c}
\toprule
\textbf{Prompt} & \textbf{Model} & \textbf{``False'' Rate} \\
\midrule

\multirow{7}{*}{Zero-shot}
 & Llama-3.1-8B-Instruct & 0.00 \\
 & Mistral-7B-Instruct-v0.3 & 0.01 \\
 & Qwen2.5-7B-Instruct & 0.00 \\
 & Yi-1.5-9B-Chat & 0.00 \\
 & deepseek-llm-7b-chat & 0.00 \\
 & gemma-2-9b-it & 0.00 \\
 & glm-4-9b-chat-hf & 0.00 \\
\midrule
\multirow{7}{*}{DAP}
 & Llama-3.1-8B-Instruct & 0.19 \\
 & Mistral-7B-Instruct-v0.3 & 0.00 \\
 & Qwen2.5-7B-Instruct & 0.02 \\
 & Yi-1.5-9B-Chat & 0.00 \\
 & deepseek-llm-7b-chat & 0.00 \\
 & gemma-2-9b-it & 0.00 \\
 & glm-4-9b-chat-hf & 0.00 \\
\midrule

\multirow{7}{*}{COT}
 & Llama-3.1-8B-Instruct & 0.00 \\
 & Mistral-7B-Instruct-v0.3 & 0.03 \\
 & Qwen2.5-7B-Instruct & 0.00 \\
 & Yi-1.5-9B-Chat & 0.02 \\
 & deepseek-llm-7b-chat & 0.00 \\
 & gemma-2-9b-it & 0.00 \\
 & glm-4-9b-chat-hf & 0.00 \\
\midrule

\multirow{7}{*}{Counterfactual}
 & Llama-3.1-8B-Instruct & 0.03 \\
 & Mistral-7B-Instruct-v0.3 & 0.02 \\
 & Qwen2.5-7B-Instruct & 0.00 \\
 & Yi-1.5-9B-Chat & 0.00 \\
 & deepseek-llm-7b-chat & 0.00 \\
 & gemma-2-9b-it & 0.00 \\
 & glm-4-9b-chat-hf & 0.07 \\

\bottomrule
\end{tabular}
\caption{“False” rate (Group C) across different prompt settings and models.}
\label{tab:false_rate_group_c}
\end{table}

\paragraph{Detailed Results by Semantic Verb Class.}
Tables \ref{tab:verbclassa}-\ref{tab:verbclasscbias_whatif} provide the detailed accuracy breakdown (Accuracy results for Group A and C) and bias breakdown (TBR results for Group C) for each model across the four semantic sub-classes of telic verbs.


\newpage
\begin{table*}[!htbp]
\centering
\renewcommand\arraystretch{0.9}
\small
\begin{tabular}{lccccc}
\toprule
Models & Change of State & Consumption & Creation & Motion to Goal \\
\midrule
Llama-3.1-8B-Instruct & 0.48 & 0.56 & 0.38 & 0.75 \\
Mistral-7B-Instruct-v0.3 & 0.32 & 0.44 & 0.31 & 0.88 \\
Qwen2.5-7B-Instruct & 0.20 & 0.11 & 0.15 & 0.50 \\
Yi-1.5-9B-Chat & 0.34 & 0.67 & 0.23 & 0.62 \\
deepseek-llm-7b-chat & 0.05 & 0.00 & 0.05 & 0.00 \\
gemma-2-9b-it & 0.02 & 0.00 & 0.05 & 0.00 \\
glm-4-9b-chat-hf & 0.09 & 0.33 & 0.08 & 0.50 \\
\bottomrule
\end{tabular}
\caption{Accuracy Breakdown by Verb Class in Zero-shot (Group A)}
\label{tab:verbclassa}
\end{table*}

\begin{table*}[!htbp]
\centering
\renewcommand\arraystretch{0.9}
\small
\begin{tabular}{lccccc}
\toprule
Models & Change of State & Consumption & Creation & Motion to Goal \\
\midrule
Llama-3.1-8B-Instruct & 0.02 & 0.00 & 0.03 & 0.00 \\
Mistral-7B-Instruct-v0.3 & 0.00 & 0.00 & 0.03 & 0.12 \\
Qwen2.5-7B-Instruct & 0.43 & 0.33 & 0.51 & 0.62 \\
Yi-1.5-9B-Chat & 0.02 & 0.11 & 0.00 & 0.00 \\
deepseek-llm-7b-chat & 0.00 & 0.00 & 0.00 & 0.00 \\
gemma-2-9b-it & 0.02 & 0.11 & 0.05 & 0.25 \\
glm-4-9b-chat-hf & 0.00 & 0.11 & 0.03 & 0.12 \\
\bottomrule
\end{tabular}
\caption{Accuracy Breakdown by Verb Class in Zero-shot (Group C)}
\label{tab:verbclassc}
\end{table*}

\begin{table*}[!htbp]
\centering
\renewcommand\arraystretch{0.9}
\small
\begin{tabular}{lcccc}
\toprule
Models & Change of State & Consumption & Creation & Motion to Goal \\
\midrule
Llama-3.1-8B-Instruct & 0.98 & 1.00 & 0.97 & 1.00 \\
Mistral-7B-Instruct-v0.3 & 1.00 & 1.00 & 0.95 & 0.88 \\
Qwen2.5-7B-Instruct & 0.57 & 0.67 & 0.49 & 0.38 \\
Yi-1.5-9B-Chat & 0.98 & 0.89 & 1.00 & 1.00 \\
deepseek-llm-7b-chat & 1.00 & 1.00 & 1.00 & 1.00 \\
gemma-2-9b-it & 0.98 & 0.89 & 0.95 & 0.75 \\
glm-4-9b-chat-hf & 1.00 & 0.89 & 0.97 & 0.88 \\
\bottomrule
\end{tabular}
\caption{Bias Breakdown by Verb Class in Zero-shot (Group C)}
\label{tab:verbclasscbias}
\end{table*}

\begin{table*}[!htbp]
\centering
\renewcommand\arraystretch{0.9}
\small
\begin{tabular}{lcccc}
\toprule
Models & Change of State & Consumption & Creation & Motion to Goal \\
\midrule
Llama-3.1-8B-Instruct  & 0.66 & 0.56 & 0.49 & 0.88 \\
Mistral-7B-Instruct-v0.3  & 0.55 & 0.67 & 0.44 & 1.00 \\
Qwen2.5-7B-Instruct  & 0.34 & 0.33 & 0.28 & 0.75 \\
Yi-1.5-9B-Chat  & 0.61 & 0.78 & 0.44 & 0.75 \\
deepseek-llm-7b-chat  & 0.11 & 0.11 & 0.15 & 0.00 \\
gemma-2-9b-it  & 0.09 & 0.00 & 0.05 & 0.25 \\
glm-4-9b-chat-hf  & 0.02 & 0.33 & 0.05 & 0.25 \\
\bottomrule
\end{tabular}
\caption{Accuracy Breakdown by Verb Class in DAP (Group A)}
\label{tab:verbclassa_dap}
\end{table*}

\begin{table*}[!htbp]
\centering
\renewcommand\arraystretch{0.9}
\small
\begin{tabular}{lcccc}
\toprule
Models & Change of State & Consumption & Creation & Motion to Goal \\
\midrule
Llama-3.1-8B-Instruct  & 0.30 & 0.11 & 0.54 & 0.12 \\
Mistral-7B-Instruct-v0.3  & 0.07 & 0.00 & 0.05 & 0.12 \\
Qwen2.5-7B-Instruct & 0.84 & 0.78 & 0.95 & 1.00 \\
Yi-1.5-9B-Chat & 0.20 & 0.22 & 0.15 & 0.38 \\
deepseek-llm-7b-chat & 0.00 & 0.00 & 0.00 & 0.00 \\
gemma-2-9b-it & 0.61 & 0.56 & 0.56 & 0.75 \\
glm-4-9b-chat-hf & 0.34 & 0.44 & 0.28 & 0.38 \\
\bottomrule
\end{tabular}
\caption{Accuracy Breakdown by Verb Class in DAP (Group C)}
\label{tab:verbclassc_dap}
\end{table*}

\begin{table*}[!htbp]
\centering
\renewcommand\arraystretch{0.9}
\small
\begin{tabular}{lcccc}
\toprule
Models & Change of State & Consumption & Creation & Motion to Goal \\
\midrule
Llama-3.1-8B-Instruct  & 0.48 & 0.78 & 0.33 & 0.50 \\
Mistral-7B-Instruct-v0.3  & 0.93 & 1.00 & 0.95 & 0.88 \\
Qwen2.5-7B-Instruct  & 0.14 & 0.22 & 0.03 & 0.00 \\
Yi-1.5-9B-Chat  & 0.80 & 0.78 & 0.85 & 0.62 \\
deepseek-llm-7b-chat  & 1.00 & 1.00 & 1.00 & 1.00 \\
gemma-2-9b-it  & 0.39 & 0.44 & 0.44 & 0.25 \\
glm-4-9b-chat-hf  & 0.66 & 0.56 & 0.72 & 0.62 \\
\bottomrule
\end{tabular}
\caption{Bias Breakdown by Verb Class in DAP (Group C)}
\label{tab:verbclasscbias_dap}
\end{table*}

\begin{table*}[!htbp]
\centering
\renewcommand\arraystretch{0.9}
\small
\begin{tabular}{lccccc}
\toprule
Models & Change of State & Consumption & Creation & Motion to Goal \\
\midrule
Llama-3.1-8B-Instruct & 0.14 & 0.22 & 0.13 & 0.25 \\
Mistral-7B-Instruct-v0.3 & 0.18 & 0.22 & 0.23 & 0.50 \\
Qwen2.5-7B-Instruct & 0.30 & 0.11 & 0.28 & 0.62 \\
Yi-1.5-9B-Chat & 0.64 & 0.67 & 0.38 & 0.88 \\
deepseek-llm-7b-chat & 0.09 & 0.00 & 0.03 & 0.00 \\
gemma-2-9b-it & 0.18 & 0.00 & 0.08 & 0.50 \\
glm-4-9b-chat-hf & 0.20 & 0.44 & 0.21 & 0.38 \\
\bottomrule
\end{tabular}
\caption{Accuracy Breakdown by Verb Class in CoT (Group A)}
\label{tab:verbclassa_cot}
\end{table*}

\begin{table*}[!htbp]
\centering
\renewcommand\arraystretch{0.9}
\small
\begin{tabular}{lccccc}
\toprule
Models & Change of State & Consumption & Creation & Motion to Goal \\
\midrule
Llama-3.1-8B-Instruct & 0.64 & 1.00 & 0.64 & 0.62 \\
Mistral-7B-Instruct-v0.3 & 0.36 & 0.78 & 0.38 & 0.38 \\
Qwen2.5-7B-Instruct & 0.93 & 1.00 & 1.00 & 1.00 \\
Yi-1.5-9B-Chat & 0.89 & 0.78 & 0.82 & 0.88 \\
deepseek-llm-7b-chat & 0.23 & 0.33 & 0.62 & 0.38 \\
gemma-2-9b-it & 1.00 & 1.00 & 0.97 & 0.88 \\
glm-4-9b-chat-hf & 0.30 & 0.44 & 0.46 & 0.38 \\
\bottomrule
\end{tabular}
\caption{Accuracy Breakdown by Verb Class in CoT (Group C)}
\label{tab:verbclassc_cot}
\end{table*}

\begin{table*}[!htbp]
\centering
\renewcommand\arraystretch{0.9}
\small
\begin{tabular}{lcccc}
\toprule
Models & Change of State & Consumption & Creation & Motion to Goal \\
\midrule
Llama-3.1-8B-Instruct & 0.36 & 0.00 & 0.36 & 0.38 \\
Mistral-7B-Instruct-v0.3 & 0.61 & 0.22 & 0.56 & 0.62 \\
Qwen2.5-7B-Instruct & 0.07 & 0.00 & 0.00 & 0.00 \\
Yi-1.5-9B-Chat & 0.09 & 0.11 & 0.18 & 0.12 \\
deepseek-llm-7b-chat & 0.77 & 0.67 & 0.38 & 0.62 \\
gemma-2-9b-it & 0.00 & 0.00 & 0.03 & 0.12 \\
glm-4-9b-chat-hf & 0.70 & 0.56 & 0.54 & 0.62 \\
\bottomrule
\end{tabular}
\caption{Bias Breakdown by Verb Class in CoT (Group C)}
\label{tab:verbclasscbias_cot}
\end{table*}

\begin{table*}[!htbp]
\centering
\renewcommand\arraystretch{0.9}
\small
\begin{tabular}{lccccc}
\toprule
Models & Change of State & Consumption & Creation & Motion to Goal \\
\midrule
Llama-3.1-8B-Instruct & 0.27 & 0.22 & 0.44 & 0.50 \\
Mistral-7B-Instruct-v0.3 & 0.30 & 0.33 & 0.15 & 0.75 \\
Qwen2.5-7B-Instruct & 0.07 & 0.11 & 0.05 & 0.12 \\
Yi-1.5-9B-Chat & 0.05 & 0.11 & 0.05 & 0.25 \\
deepseek-llm-7b-chat & 0.23 & 0.44 & 0.26 & 0.12 \\
gemma-2-9b-it & 0.02 & 0.00 & 0.05 & 0.00 \\
glm-4-9b-chat-hf & 0.25 & 0.44 & 0.08 & 0.62 \\
\bottomrule
\end{tabular}
\caption{Accuracy Breakdown by Verb Class in Counterfactual (Group A)}
\label{tab:verbclassa_whatif}
\end{table*}

\begin{table*}[!htbp]
\centering
\renewcommand\arraystretch{0.9}
\small
\begin{tabular}{lccccc}
\toprule
Models & Change of State & Consumption & Creation & Motion to Goal \\
\midrule
Llama-3.1-8B-Instruct  & 1.00 & 0.89 & 1.00 & 0.75 \\
Mistral-7B-Instruct-v0.3  & 0.89 & 0.67 & 0.90 & 0.88 \\
Qwen2.5-7B-Instruct & 1.00 & 1.00 & 1.00 & 1.00 \\
Yi-1.5-9B-Chat & 1.00 & 1.00 & 1.00 & 1.00 \\
deepseek-llm-7b-chat & 0.00 & 0.00 & 0.00 & 0.00 \\
gemma-2-9b-it & 1.00 & 1.00 & 1.00 & 1.00 \\
glm-4-9b-chat-hf  & 0.70 & 0.67 & 0.79 & 0.62 \\
\bottomrule
\end{tabular}
\caption{Accuracy Breakdown by Verb Class in Counterfactual (Group C)}
\label{tab:verbclassc_whatif}
\end{table*}

\begin{table*}[!htbp]
\centering
\renewcommand\arraystretch{0.9}
\small
\begin{tabular}{lcccc}
\toprule
Models & Change of State & Consumption & Creation & Motion to Goal \\
\midrule
Llama-3.1-8B-Instruct & 0.00 & 0.00 & 0.00 & 0.00 \\
Mistral-7B-Instruct-v0.3 & 0.11 & 0.33 & 0.05 & 0.12 \\
Qwen2.5-7B-Instruct & 0.00 & 0.00 & 0.00 & 0.00 \\
Yi-1.5-9B-Chat & 0.00 & 0.00 & 0.00 & 0.00 \\
deepseek-llm-7b-chat & 1.00 & 1.00 & 1.00 & 1.00 \\
gemma-2-9b-it & 0.00 & 0.00 & 0.00 & 0.00 \\
glm-4-9b-chat-hf & 0.20 & 0.33 & 0.15 & 0.25 \\
\bottomrule
\end{tabular}
\caption{Bias Breakdown by Verb Class in Counterfactual (Group C)}
\label{tab:verbclasscbias_whatif}
\end{table*}

\end{document}